%% file: root.tex
\documentclass[preprint,5p,times,twocolumn]{elsarticle}

\usepackage{lineno,hyperref}
\usepackage{graphics} 
\usepackage[caption=false,font=footnotesize]{subfi
g}
\usepackage{epsfig} 
\usepackage{xcolor}
\DeclareMathAlphabet{\pazocal}{OMS}{zplm}{m}{n}
\usepackage{amsmath} 
\usepackage{amssymb}  
\usepackage{threeparttable}
\usepackage{booktabs}

\usepackage{url}
\usepackage{hyperref}
\usepackage{color}
\usepackage{picins}
\usepackage{lipsum}  

\newcommand\blfootnote[1]{%
  \begingroup
  \renewcommand\thefootnote{}\footnote{#1}%
  \addtocounter{footnote}{-1}%
  \endgroup
}

\journal{Journal of Systems Architecture}

\begin{document}

\begin{frontmatter}


\title{A Software Architecture for Autonomous Vehicles: Team LRM-B Entry in the First CARLA Autonomous Driving Challenge\tnoteref{mytitlenote}}
\tnotetext[mytitlenote]{This study was financed in part by the Coordena\c{c}\~ao de Aperfei\c{c}oamento de Pessoal de N\'ivel Superior - Brasil (CAPES) - Finance Code 001 and grants 88882.328851/2019-01, 88887.136349/2017-00 and 88882.328835/2019-01; the Brazilian National Research Council (CNPq) under grants 132756/2018-8 and 465755/2014-3; and the S\~ao Paulo Research Foundation (FAPESP) under grants 2019/27301-7, 2018/19732-5, 2019/03366-2 and  2014/50851-0.}

\author[icmcaddress]{Luis Alberto Rosero\corref{cor1}}
\ead{lrosero@usp.br}
\author[icmcaddress]{Iago Pacheco Gomes}
\ead{iagogomes@usp.br}
\author[icmcaddress]{J\'unior Anderson Rodrigues da Silva}
\ead{junior.anderson@usp.br}
\author[icmcaddress]{Tiago Cesar dos Santos}
\ead{tiagocs@icmc.usp.br}
\author[icmcaddress]{Angelica Tiemi Mizuno Nakamura}
\ead{tiemi.mizuno@usp.br}
\author[icmcaddress]{Jean Amaro}
\ead{jean.amaro@usp.br}
\author[icmcaddress]{Denis Fernando Wolf}
\ead{denis@icmc.usp.br}
\author[icmcaddress]{Fernando Santos Os\'orio}
\ead{fosorio@usp.br}

\cortext[cor1]{Corresponding author}
\address[icmcaddress]{Institute of Mathematics and Computer Science, University of S\~ao Paulo, Av Trabalhador S\~ao-carlense, 400\\
S\~ao Carlos, Brazil}

\begin{abstract}

The objective of the first CARLA autonomous driving challenge was to deploy autonomous driving systems to lead with complex traffic scenarios where all participants faced the same challenging traffic situations. According to the organizers, this competition emerges as a way to democratize and to accelerate the research and development of autonomous vehicles around the world using the CARLA simulator contributing to the development of the autonomous vehicle area. Therefore, this paper presents the architecture design for the navigation of an autonomous vehicle in a simulated urban environment that attempts to commit the least number of traffic infractions, which used as the baseline the original architecture of the platform for autonomous navigation CaRINA 2. 
Our agent traveled in simulated scenarios for several hours, demonstrating his capabilities, winning three out of the four tracks of the challenge, and being ranked second in the remaining track. 
 Our architecture was made towards meeting the requirements of CARLA Autonomous Driving Challenge and has components for obstacle detection using 3D point clouds, traffic signs detection and classification which employs Convolutional Neural Networks (CNN) and depth information, risk assessment with collision detection using short-term motion prediction, decision-making with Markov Decision Process (MDP), and control using Model Predictive Control (MPC).

\end{abstract}

\begin{keyword}
\texttt CARLA Simulator\sep Software Architecture\sep Autonomous Vehicles\sep Obstacle Detection\sep Traffic Sign Detection\sep Decision Making
\end{keyword}
 
\end{frontmatter}
\blfootnote{\copyright 2020. This manuscript version is made available under the CC-BY-NC-ND 4.0 license \url{http://creativecommons.org/licenses/by-nc-nd/4.0/}} 

\section{Introduction}
\label{sec:introduction}
\input{tex/introduction.tex}

\section{CARLA Autonomous Driving Challenge}
\label{sec:cadc}
\input{tex/CADCH.tex}

\section{Related Work}
\label{sec:related_works}
\input{tex/related_works.tex}

\section{Software Architecture}
\label{sec:sw_arch}
\input{tex/architecture.tex}

\section{RESULTS AND DISCUSSION}
\label{sec:results}
\input{tex/results.tex}

\section{CONCLUSIONS}
\label{sec:conclusion}
\input{tex/conclusion.tex}


\bibliography{root}

\end{document}

%% file: tex/introduction.tex
Autonomous Vehicles are intelligent and robotic vehicles that navigate in traffic without human intervention, dealing with all driving scenarios and respecting traffic legislation \cite{bimbraw2015autonomous}. They combine a wide range of sensors and software components towards creating a rich representation of a scene to understand the external environment, make decisions, and take actions similarly to, or even better than human drivers \cite{liu2017creating}. Therefore, they are expected to transform urban traffic improving mobility, safety and accessibility, reducing pollutant emissions, among other benefits offered \cite{fagnant2015preparing}. 

All workflow of an autonomous system is performed by various software components and algorithms, which use concepts of machine learning, computer vision, decision theory, probability theory, control theory, and other research fields. Such a heterogeneous characteristic increases the complexity of its development and evaluation. According to Kalra and Paddock  \cite{kalra2016driving}, and Koopman and Wagner \cite{koopman2016challenges}, to reliably assess an autonomous system, more than road tests are required, due to the complexity and variability of the driving scenario and number of components of the system. 

Autonomous driving challenges have been one of the major factors that contribute to the development and advance in the research field since \emph{DARPA's Grand Challenge} (2004 and 2005) \cite{buehler20072005} and \emph{DARPA's Urban Challenge} (2007)  \cite{buehler2009darpa}. Other challenges, such as \emph{Intelligent Vehicle Future Challenge} (IVFC), organized by the National Natural Science Foundation of China (NSFC) \cite{wang2017ai}, \emph{Autonomous Vehicle Challenge} (AVC), sponsored by Hyundai Motor Group in South Korea \cite{jo2015development}, and \emph{European Truck Platooning Challenge} (EU TPC), promoted by The European Commission \cite{aarts2016european} have also been important for advances in the autonomous vehicle research field.

Although those challenges significantly impact the progress of the research field, the performance evaluation of autonomous system is limited to some particular situations created specifically for competitions or common traffic, since the primary concern in real-world tests is participants' safety, thus limiting the tests to more controlled situations. 

According to Huang et al. \cite{huang2016autonomous}, extensive evaluations of autonomous systems require simulation, which can evaluate single components or the complete system. One of the primary advantages of simulators is they can create specific scenarios to be repeated as many times as necessary for assessments of the performance of a component, as well as several driving situations with realistic dynamics and environmental feedback, such as weather conditions, sensor malfunctions, traffic violations, hazardous traffic events, traffic jams, crowded streets, among others. Moreover, simulation platforms can benchmark autonomous driving systems, since they can evaluate different systems approaches with the same set of events and conditions. 

Therefore, the CARLA (CAR Learning to Act) simulator staff promoted \emph{CARLA Autonomous Driving Challenge} (CARLA AD Challenge) competition, in 2019, which comprised two categories and four tracks. Its major purpose was to evaluate safe driving task for autonomous systems developed by different worldwide research groups, which could also rely on different sets of sensors available on each track. 

This paper introduces the software architecture designed by the LRM-B team, who participated in the CARLA AD Challenge, won three tracks and was ranked second on the remaining track. \emph{CaRINA 2} (Intelligent Robotic Car for Autonomous Navigation) software architecture was used as a baseline for the architecture; it is an autonomous vehicle research platform developed by the Mobile Robotics Lab (LRM) at the University of S\~ao Paulo Paulo, Brazil \cite{Fernandes2014}, which conducts research on intelligent and autonomous vehicles technologies. 

The main contributions of our study include: 

\begin{itemize}
    \item a modular software architecture for autonomous vehicles that works with different set of sensors; 
    
    \item a risk assessment that uses short-term motion prediction for collision detection and decision-making; 
    
    \item a probabilistic decision-making approach for a high-level longitudinal control of vehicle that handles obstacles and signalized intersections; and 
   
   \item an obstacle detector based on sensor fusion that uses deep learning for image object detection and alignment with segmented objects into a 3D point cloud (stereo and LiDAR). 
    
\end{itemize}

The remainder of the paper is organized as follows: Section \ref{sec:cadc} provides an overview of CARLA AD Challenge; Section \ref{sec:related_works} highlights the concepts and components of a general autonomous vehicle software architecture related to this research; Section \ref{sec:sw_arch} describes the software architecture developed by the LRM-B team in the competition; Section \ref{sec:results} discusses the results of the competition and other experiments; finally, Section \ref{sec:conclusion} addresses the final remarks and suggests some future work. 

%% file: tex/CADCH.tex
\begin{figure}[!tb]
	\centering
    \subfloat[Negotiation at roundabout]{
		\includegraphics[width=1.6in]{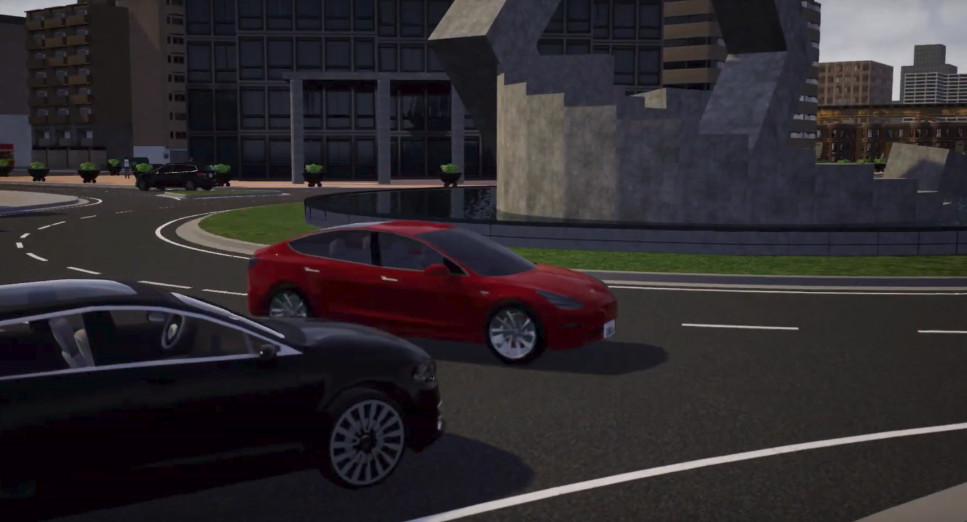}
	}
	\hfil
	\subfloat[Obstacle Avoidance]{
		\includegraphics[width=1.4in, height=.87in]{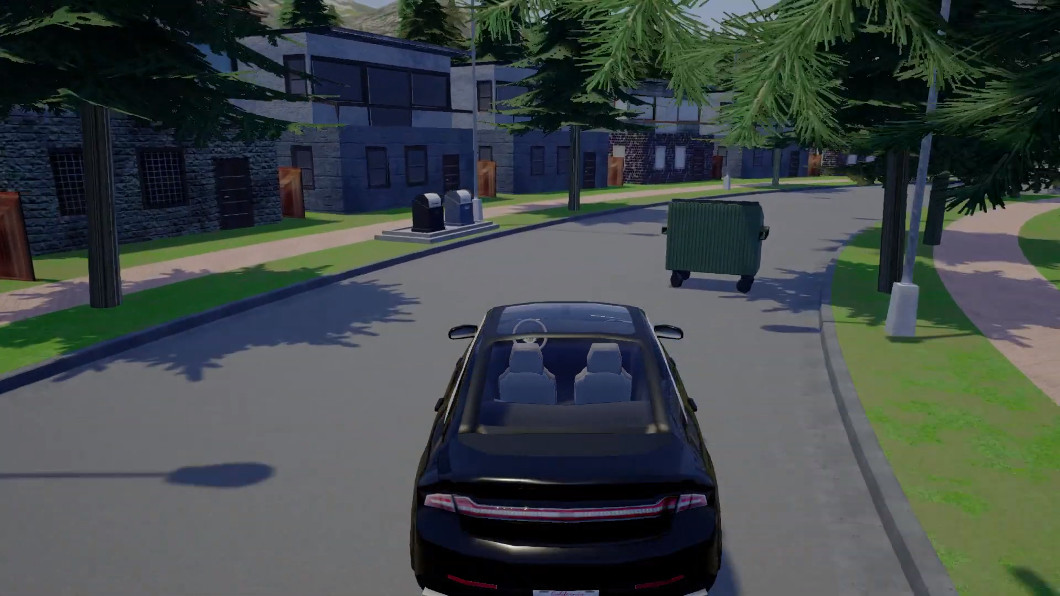}
	}
	\hfil
	\subfloat[Traffic Jam]{
		\includegraphics[width=1.6in, height=.87in]{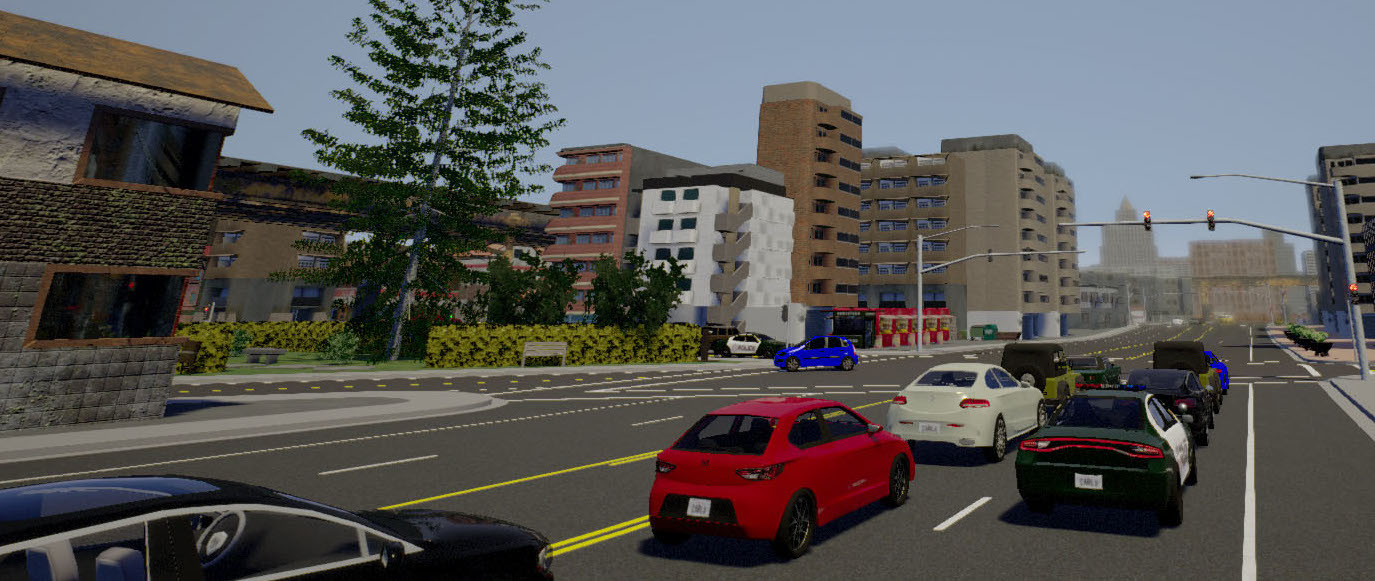}
	}
	\hfil
	\subfloat[Highways]{
		\includegraphics[width=1.4in, height=.87in]{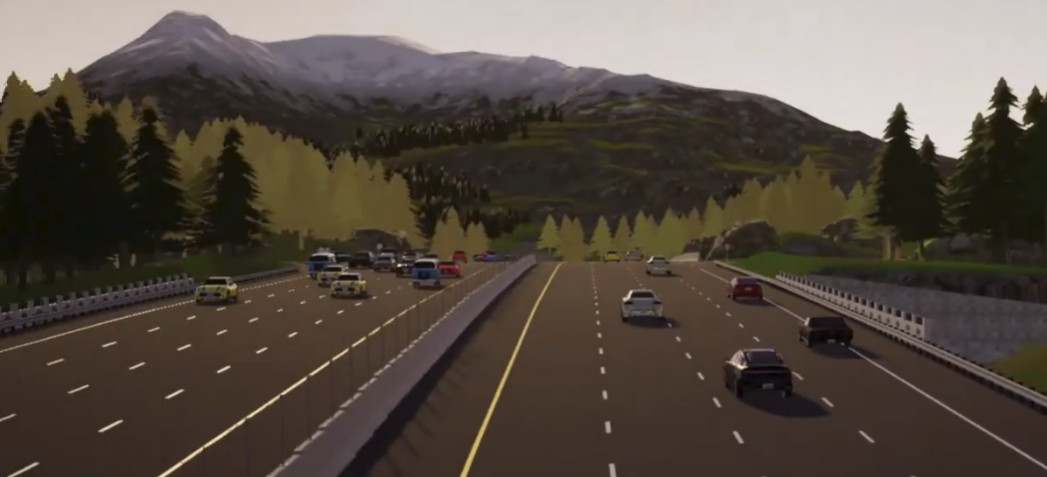}
	}
	\caption{CARLA}
	\label{fig:carla_scenes}
\end{figure}

The first CARLA Autonomous Driving Challenge (CADC) \cite{carlachallenge} was a competition for autonomous vehicles in urban environments that used CARLA (CAR Learning to Act) open source simulator, which supports the development, training and validation of several architectures and approaches for autonomous driving. CARLA provides free access to many digital features such as urban layout, buildings, traffic signs, vehicles, and pedestrians, supports various weather conditions (e.g. sunny day, fog, heavy rain), sensors (e.g. cameras, LiDAR, GPS), as well as simulations of many specific driving scenarios \cite{Dosovitskiy2017}. Fig. \ref{fig:carla_scenes} shows some of the scenarios of the competition. 

The major purpose of the competition was to evaluate safe driving task for autonomous systems and settle a benchmark of different approaches of worldwide systems. It was divided into two categories, namely \emph{Perception Heavy} and \emph{Map-based}, which were also divided into two tracks differentiated by the set of sensors available for each system design. 

The main concern of \emph{Perception Heavy} (\emph{track1} and \emph{track2}) was the development of perception algorithms for modeling and understanding the surroundings and, therefore, supplying the navigation stack (composed of components for decision making, path planning, and motion planning). The sensors available for \emph{track1} were LiDAR, GPS and Cameras. In contrast, only cameras and GPS were available for \emph{track2}. Both tracks were provided with a topological sparse representation of the route at the beginning of each route.  

Tracks in \emph{Map-based} category (\emph{track3} and \emph{track4}) provided prior knowledge of the environment for the autonomous system. In \emph{track3}, the sensors available were LiDAR, GPS, Cameras, High Definition Map that used 3D point cloud, and a set of waypoints in a geodesic coordinate system representing the route. \emph{Track4} provided all perception information (e.g. position of obstacles and traffic signs, status of traffic lights, and waypoints of lanes along the route), being the navigation stack the main concern for this track.

The evaluation metric of the competition relied on the traffic rules compliance of the autonomous system along the routes within a time limit (estimated from the route length), in which each traffic violation penalized the score, and the stretches of completed route added bonus points to the score. Different driving scenarios were used in each route, according to traffic scenarios selected from the NHTSA pre-crash typology \cite{carlachallenge}, such as control loss with no previous action, avoidance of unexpected obstacles, and crossing negotiation at an unsignalized intersection. Their simulation enabled assessments of the performance of each autonomous system in hazardous and challenging traffic scenarios with which the autonomous vehicle should deal in real-world traffic.  

This article presents the overall software architecture design for all tracks, developed by LRM-B team, winner of \emph{track1}, \emph{track3}, and \emph{track4}, and ranked second on \emph{track2} \footnote{\url{https://carlachallenge.org/results-challenge-2019/}}.  Although each track comprised a different set of sensors, the architecture provides a general interface between all components for all tracks, which enabled the team to take advantage of the features of layered architectures, such as modularity, reusability and maintainability. 

%% file: tex/related_works.tex
The layered architecture pattern establishes a topological and hierarchical organization of
the software components.  Each layer gathers software components with similar features and abstraction level. The communication between components can be classified into intra-layer and inter-layer. The former means the communication of components belonging to the same layer, while the latter between components of different layers. Both communication styles use a well-defined communication interface, which allows the replacement of specific components or the entire layer, keeping the system behavior. This property reinforces the modularity, scalability, reusability, and maintainability of the architectural style \cite{zhu2005software}\cite{chavan2015review}. This architecture style is suitable to design autonomous systems since the workflow of the system is carried out by many components which have correlated meaning and can be arranged hierarchically \cite{chavan2015review}.  

The layered architecture pattern establishes a topological and hierarchical organization of the software components. Each layer gathers software components with similar features and abstraction level. The communication between components can be classified into intra-layer and inter-layer. The former refers to communication of components that belong to the same layer, whereas the latter denotes communication between components of different layers. Both communication styles use a well-defined communication interface, which enables the replacement of specific components or the entire layer, keeping the system's behavior. Such a property reinforces the modularity, scalability, reusability, and maintainability of the architectural style \cite{zhu2005software}\cite{chavan2015review}, which is suitable for the design of autonomous systems, since the workflow is carried out  by several components with correlated functions and can be arranged hierarchically \cite{chavan2015review}. 

According to Liu et al. \cite{liu2017creating} and Kato et al. \cite{kato2015open}, the standard layers of an autonomous system can be categorized as sensing, perception, path planning, decision-making, control, system-management, and human-vehicle interface. The \emph{sensing layer} makes data from sensors available to other components of the system, which is in line with the architecture design of some autonomous vehicles such as \emph{A1} \cite{jo2015development}, \emph{BerthaOne} \cite{tacs2018making},  and  \emph{AutoWare} \cite{kato2018autoware}. 

The components of the \emph{perception layer} use those data to obtain useful information, e.g., position of pedestrians, and vehicles or objects in the image. Another component of \emph{perception layer} is localization, which estimates the position of the vehicle in a specific coordinate system (e.g., global or local) \cite{kuutti2018survey}.   Wei et al.  \cite{wei2013towards} used a filter that integrated GPS (Global Position System) with RTK (Real Time Kinematic) correction, wheel speed sensors, and IMU (Inertial Measurement Unit) for global localization systems with centimeter-level precision. Alternatively, Kato et al. \cite{kato2018autoware} performed localization by scan matching between 3D maps and LiDAR (Light Detection And Ranging) scanners. The perception of both vehicles was also responsible for tasks such as obstacle detection, traffic signs detection, and traffic lights detection using algorithms based on image processing (e.g. Convolution Neural Networks \cite{yolov3,he2017mask, ren2015faster}) and point cloud clustering \cite{rosero2017calibration, zermas2017fast,ye2016object,kellner2012grid,kato2015open}. 

The information from the \emph{perception layer} is taken into account by the \emph{path planning layer} and \emph{decision-making layer}, which must find a path from the current position of the vehicle towards its destination, and decide on the behavior of the vehicle according to both driving situations and traffic laws. Jo et al. \cite{jo2015development} and Wei et al. \cite{wei2014behavioral} claimed such layers can be divided into three stages, namely global planning, behavior generation, and motion planning. The former, also known as route planning, computes the overall route between the current position to the destination using road-network models (e.g. High-Definition Maps \cite{li2017lane} and Lane-level Annotated Maps \cite{poggenhans2018lanelet2}\cite{jiang2019flexible}) and heuristic functions (e.g. shortest path, speed limits along the route, and traffic flow). 

Behavior generation, also known as behavior reasoning or decision-making, makes tactical decisions on the actions of the vehicle, e.g., maneuvers to be performed (e.g. overtaking, lane-change, lane following, U-turn or emergency-stop), or the way the vehicles should interact with other traffic participants \cite{paden2016survey,katrakazas2015real}. As stated in Paden et al. \cite{paden2016survey} and Katrakazas et al. \cite{katrakazas2015real}, some of the algorithms used for the decision process are MDP (Markov Decision Process) \cite{brechtel2011probabilistic}, POMDP (Partially-Observable Markov Decision Process) \cite{Kurniawati2016},  State Machines (e.g. Finite State Machines, or Hierarchical Concurrent State Machines) \cite{jo2015development}\cite{tacs2018making}, Behavior-Tree \cite{olsson2016behavior}, and Game Theory \cite{li2017game}.   

Based on the global plan and reasonable decision made by previous components, motion planning creates local, obstacle-free and dynamic feasible paths to be tracked by low-level controllers in the \emph{control layer}, which generate brake, throttle and steering angle commands. According to Liu et al. \cite{liu2017creating} and Massera Filho et al. \cite{massera2014longitudinal}, a way to accomplish this task is to divide the movement into lateral and longitudinal controllers, which calculate actions considering kinematic and dynamic constraints of the movement. In brief, the lateral controller uses Ackermann geometry equations to calculate the steering angle, and the longitudinal controller uses a speed profile of the path to calculate the acceleration and braking required so that the desired speed can be reached. Model Predictive Control (MPC) is a well-known technique applied for this task; it optimizes a cost function based on defined constraints and the prediction of the system behavior using its dynamic model within a time horizon \cite{obayashi2016appropriate}. 

The \emph{system-management layer} monitors all components of the system, detects and identifies faults or abnormal behaviors, and launches recovery protocols in case of faults and unexpected conditions. According to Jo et al. \cite{jo2015development}, when a fault or abnormal condition is detected, the system switches the operation mode of the vehicle to pause, so that a human driver takes control. Wei et al. \cite{wei2013towards} also used the switch of operation mode as a recovery protocol, and redundancy of hardware components, which can migrate running algorithms from the faulty component to another healthy component.  

Finally, the \emph{human-vehicle interface layer} provides the graphical tools that access the system, visualizing the components information and feedback, and also request specific missions, such as destination to be reached by the vehicle. 

%% file: tex/architecture.tex
As addressed elsewhere, an autonomous system requires several components and its architectural design provides an abstract view of the system operation and organization. In a layered architecture, the components have public and well-defined communication interfaces through which they exchange information with other components. This characteristic enables the definition of a common architecture for all tracks in this challenge, through adjustments of few components for the maintenance of the same communication interface. This strategy reduces the time spent on the development of the agents, and enables the evaluation of the autonomous navigation performance with different algorithms for a specific task.   

\begin{figure}[!tb]
      \centering
      \includegraphics[scale=0.65]{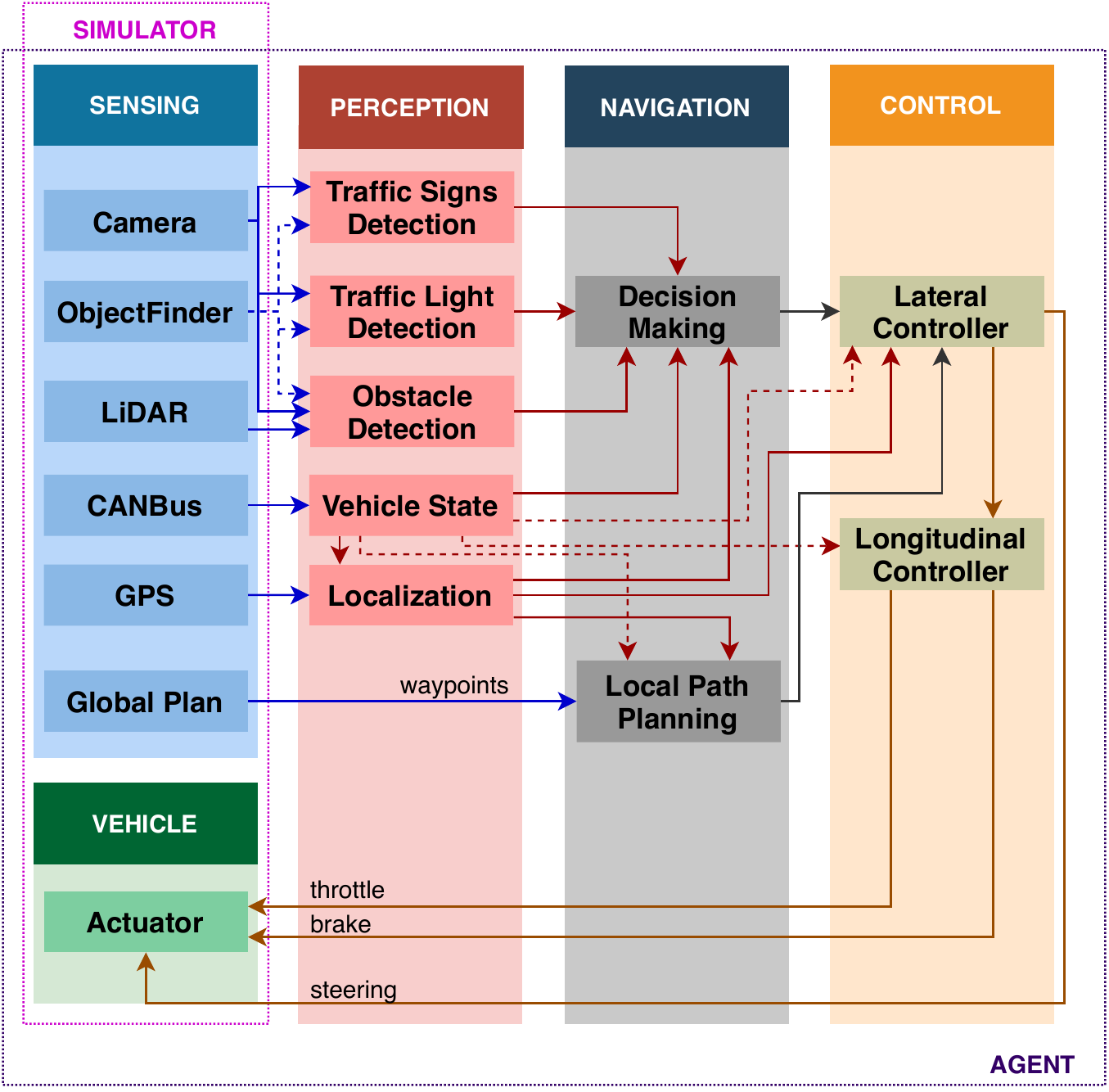}
      \caption{CaRINA Agent Software Architecture}
      \label{fig:sw_carina}
  \end{figure}

Fig. \ref{fig:sw_carina} shows the general software architecture designed for all agents of LRM-B team in the CARLA AD Challenge 2019. The name \emph{"CaRINA Agent"} is used to refer to this architecture in the rest of the article. The layers of the architecture are \emph{sensing}, \emph{perception}, \emph{navigation}, \emph{control}, and, \emph{vehicle}. Robotic framework ROS (Robotic Operating System) supported the communication interface between components with Publish/Subscribe pattern for messages passing. 

\subsection{Publish and Subscribe Model}
\input{tex/ros}

\subsection{Sensing}
\input{tex/sensing}

\subsection{Perception}
\input{tex/perception}

\subsection{Navigation}
\label{subsec:navigation}
\input{tex/navigation}

\subsection{Control}
\input{tex/control}

%% file: tex/ros.tex
The main idea of the Publisher-Subscriber style is each component can assume two roles, i.e., publisher and subscriber. The former produces some information or generates data, which are required and used by components that assume the second role. Therefore, when the publisher modifies the shared resources, it notifies all components that have requested that information \cite{zhu2005software}. 

ROS has a centralized process, called \emph{master}, which indexes each new component started, which can be both publisher and subscriber. Topics and types address the messages between publishers and subscribers. After a subscriber requests a topic, the message is directly exchanged between nodes through a peer-to-peer protocol. Therefore, the communication structure provided by the framework facilitates the development of systems with many components, such as autonomous ones \cite{quigley2009ros}. 

%% file: tex/sensing.tex
The \emph{sensing layer} comprises a set of sensors classified as exteroceptive and proprioceptive that measure, respectively, data from the external environment and the vehicle itself \cite{solutions2007robotics}. The \emph{sensing layer} components obtain raw data from the sensors to be interpreted and converted into useful information by the \emph{perception layer}, so that the system creates a representation of the vehicle state and its surroundings with some degree of reliability. 

A layer for sensor management is an important resource for architecture design; it improves both flexibility and scalability, enables the replacement of one of its components, or even the entire layer, and keeps the architecture working, as long as its communication interface remains. If the set of sensors is changed, the modification in the architecture design is minimal or even non-existent. 

\begin{figure}[!tb]
      \centering
      \includegraphics[scale=0.26]{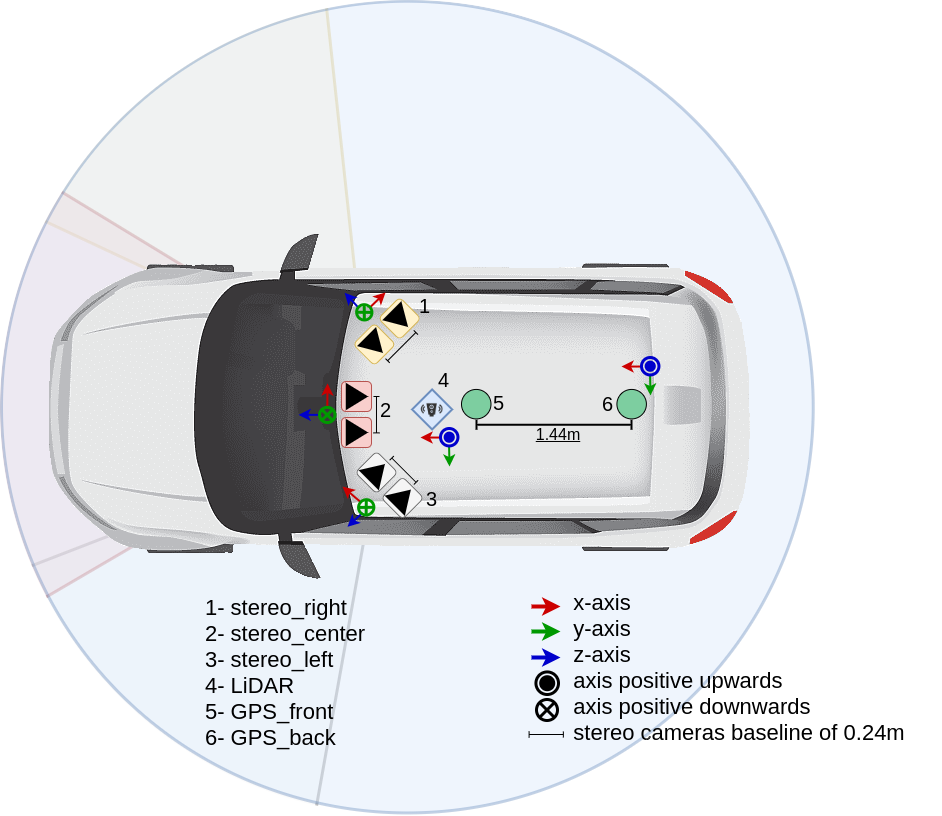}
      \caption{The sensors on the vehicle}
      \label{fig:car_sensing}
  \end{figure}

Fig. \ref{fig:car_sensing} shows the sensors layout of \emph{CaRINA Agent} for all tracks in the competition, with three pair of cameras for stereo vision, one LiDAR, and two GPS receivers. As addressed elsewhere, each track allows only one specific set of sensors, thereby, \emph{CaRINA Agent} uses only a subset of the sensors (Fig. \ref{fig:car_sensing}) in each track.  Some pseudo sensors (e.g. CANBus and ObjectFinder) are also available in the competition that provide additional information. Table \ref{tab:sensors} shows the sensors used in each track. 
\begin{table}[!tb]
\caption{Sensor}
\label{tab:sensors}
\begin{center}
\begin{tabular}{cc}
\hline
\textbf{Track}  & \textbf{Sensor}                                      \\ 
\hline
\textit{track1} & \begin{tabular}[c]{@{}c@{}}GPS; LiDAR;  stereo\_center; CANBus\end{tabular}                       \\ \hline
\textit{track2}          & \begin{tabular}[c]{@{}c@{}}GPS; stereo\_left;  stereo\_center; stereo\_right; CANBus\end{tabular} \\ \hline
\textit{track3}          & \begin{tabular}[c]{@{}c@{}}GPS; LiDAR;  stereo\_center; CANBus\end{tabular}                       \\ \hline
\textit{track4}          & \begin{tabular}[c]{@{}c@{}}GPS; CANBus;  ObjectFinder\end{tabular}                              \\ \hline
\end{tabular}
\end{center}
\end{table}

\begin{itemize}
    \item \textbf{GPS}: This sensor provides the position of the vehicle in a geodesic coordinate frame at 10Hz frequency, defined as:
    \begin{equation}\label{eq:position}
	    p_{k}^{geo} = \left(lat_k, lon_k, alt_k\right)
    \end{equation}
    where $lat$ and $lon$  correspond to the latitude and longitude, $alt$ denotes altitude, and $k$ is the timestamp of the frame. 
    
    \item \textbf{Cameras}:
    Six cameras grouped into 3 stereo pairs placed at a height of 1.8m were used (see Fig. \ref{fig:car_sensing}). The size of the images for the stereo cameras named \emph{"Stereo\_right"} and \emph{"Stereo\_left"}, is $600\times320$ (width $\times$ height), and   for the \emph{"Stereo\_center"} camera is $1080\times540$ (width $\times$ height). The next section provides detailed information about the camera system.  
    
    \item \textbf{LiDAR}: This sensor uses $32$ laser simulated channels with $45$ degrees of vertical field of view ($15^{\circ}$ of upper fov, and $-30^{\circ}$ of lower fov), and delivers approximately $500000$ points per second, known as point cloud, with $360$ degrees horizontal field of view at $20Hz$ frequency. Each point is defined by position (x, y, z) in a Cartesian coordinate system, relative to the sensor position. The coverage of the point cloud has $50m$ radii.
    
    \item \textbf{CANBus}: This proprioceptive pseudo sensor provides information on the internal state of the vehicle, such as speed, steering angle, shape dimensions, among others.  
    
    \item \textbf{ObjectFinder}: This pseudo sensor provides the position and orientation of all dynamic (e.g. vehicles and walkers) and static (e.g. traffic signs and traffic lights) objects belonging to the simulated environment, as well as information about the state of traffic lights and shape of the objects.
   
\end{itemize}

%% file: tex/perception.tex
The components of the \emph{perception layer} transform sensor data into useful information using algorithms and theoretical definitions from different research fields, such as Computer Vision, Linear Algebra, Probability, and Machine Learning \cite{liu2017creating}. This information enables the representation of the vehicle state and environment surroundings towards supplying the components of the \emph{navigation layer} in the architectural design shown in Fig. \ref{fig:sw_carina}. 

\subsubsection{Localization}
After the position has been converted from geodesic to UTM (Universal Transverse Mercator) coordinate system ($p^{UTM}$), which is a Cartesian coordinate system for localization on Earth, the heading of the vehicle can be estimated by either two sequential timestamps ($k$ and $k+1$), or the readings of two receivers ($GPS\_front$ and $GPS\_back$) in the same timestamp. In both cases, the $arctan$ function is applied as follows:

    \begin{equation}\label{eq:heading}
        \theta = arctan\left(\frac{y_{k+1} - y_k}{x_{k+1} - x_{k}}\right)
    \end{equation}
    where $\theta$ is the heading, tuple $\left(x, y\right)$ is the position in UTM,and $k+1$ and $k$ are $front$ and $back$, respectively.  
    
    The GPS output is defined as 
    \begin{equation}\label{eq:pose}
        p = \left(x_{back}, y_{back}, alt, \theta\right)
    \end{equation}
     where $p$ is the position's vector state, $\left(x_{back}, y_{back}\right)$ are the lateral and longitudinal Cartesian coordinates from the $GPS_{back}$ (see Fig. \ref{fig:car_sensing}), or from the GPS reading at timestamp $k+1$,  $alt$  is the altitude from the sensor, and $\theta$ is the vehicle's orientation (see Eq. \ref{eq:heading}).    
  
\begin{figure}[!tb]
      \centering
      \includegraphics[scale=0.45]{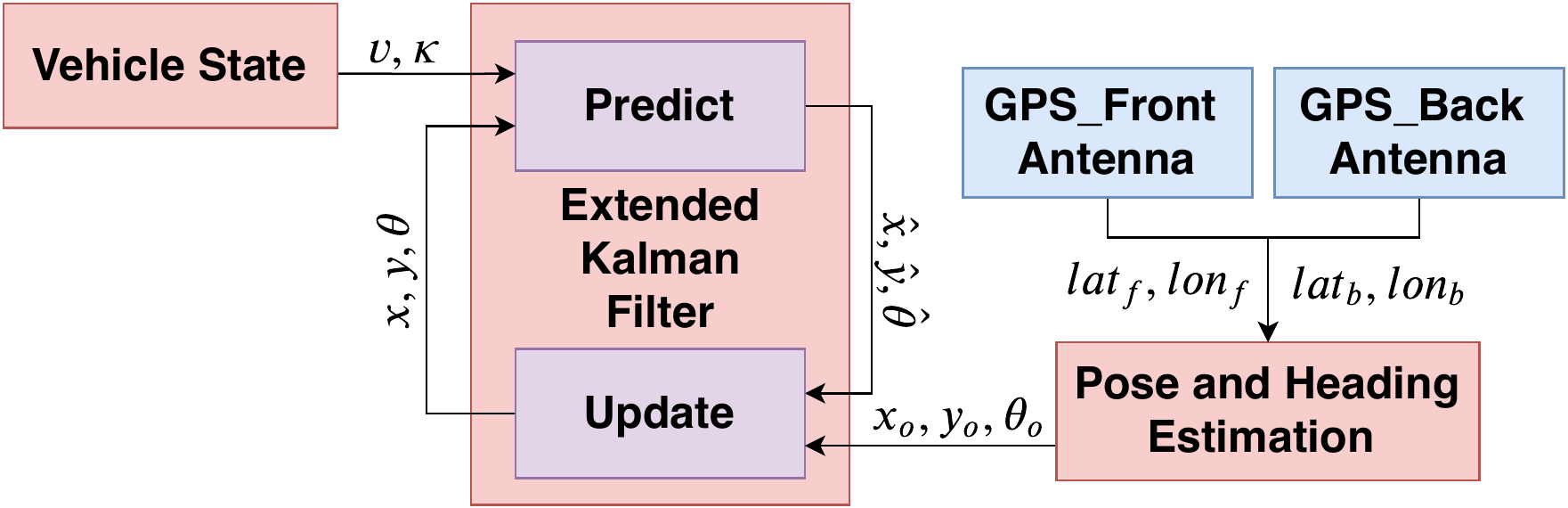}
      \caption{Flowchart of the localization system using EKF}
      \label{fig:loc_diagram}
  \end{figure}

An Extended Kalman Filter (EKF) \cite{ribeiro2004kalman} was also applied towards reducing the noise effect and avoiding a large innovation between sequential readings, which can be a result of either sensor's faults, or surrounding effects typical of GNSS-based (Global Navigation Satellites System) localization systems, such as multi-path, signal loss and antenna masking \cite{van2012sensor, bijjahalli2017gnss}.  Fig. \ref{fig:loc_diagram} displays a flowchart of the localization system using the EKF filter. 

Observation vector $z$ (Eq. \ref{eq:z_ekf}) from the GPS output $p$ (Eq. \ref{eq:pose}), a control input vector $u$ (Eq. \ref{eq:u_ekf}) from the CAN sensor, and a state vector $\hat q$ (Eq. \ref{eq:q_ekf}), which estimates the position and orientation of the vehicle, were used for \emph{CaRINA Agent}. Equation 7 shows the system's dynamic equations. 
  
  \begin{eqnarray}
  \label{eq:z_ekf}z_{k}&=&\left(x^{o}_{k}, y^{o}_k, \theta^{o}_k\right)\\
  \label{eq:u_ekf}u_{k}&=&\left(v_k, \kappa_k\right)\\
   \label{eq:q_ekf}\hat q_{k} &=& \left(\hat x_{k}, \hat y_{k}, \hat \theta_{k}\right)
  \end{eqnarray}
where, $\left(x^{o}_k, y^{o}_k, \theta^{o}_k\right)$ is the observation vector from the GPS sensor equivalent to $p=\left(x_{back}, y_{back}, \theta\right)$ (Eq. \ref{eq:pose}), $\left(v_k, \kappa_k\right)$ is the velocity and curvature from the CANBus sensor. The curvature is estimated using vehicle baseline, $L=2.85m$, and the current steering angle. In addition, $\left(\hat x_{k}, \hat y_{k}, \hat \theta_{k}\right)$ is the final pose estimation, and $k$ the timestamp .
      \begin{eqnarray}
           \label{eq:ekf_state}
           f\left(\hat q_{k-1}, u_{k}\right) &=& \begin{cases}
                                                        \hat x_{k} = \hat x_{k-1} + v_{k}\cos\left(\hat\theta_{k}\right)\Delta t      \\
                                                        \label{eq:y_ekf}\hat y_{k} = \hat y_{k-1} + v_{k}\sin\left(\hat\theta_{k}\right)\Delta t\\ 
                                                         \label{eq:t_ekf}\hat \theta_{k} = \hat \theta_{k-1} + \kappa_{k}v_{k}\Delta t
                                                \end{cases}
       \end{eqnarray}
       in which $\Delta t$ is the time elapsed between predictions. 
       
\subsubsection{Obstacle Detection}
\begin{figure}[!tb]
      \centering
      \includegraphics[scale=0.7]{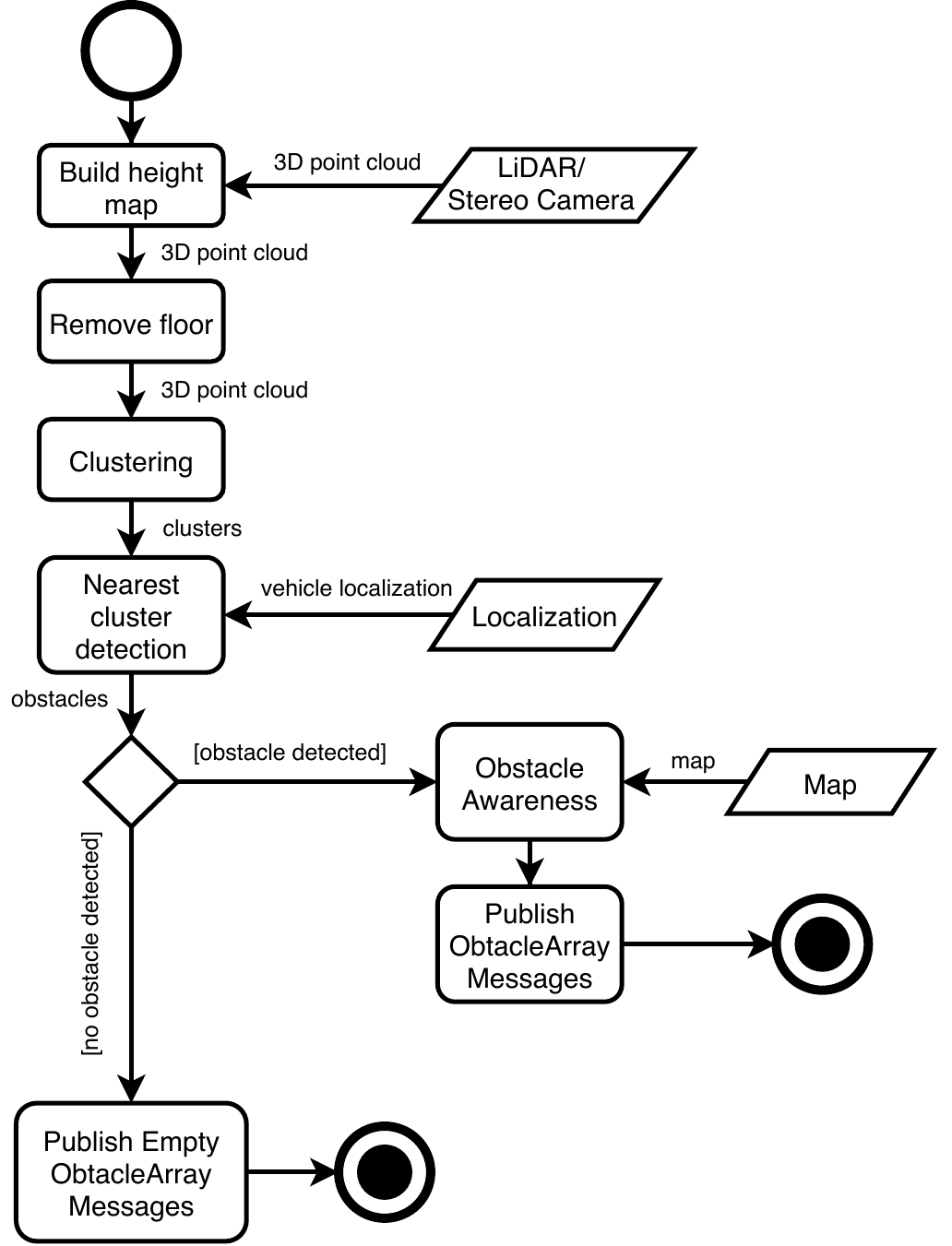}
      \caption{Objection Detection Flowchart}
      \label{fig:obst_detection}
  \end{figure}

\emph{CaRINA Agent} employs two vision systems, namely LiDAR-based and stereo camera-based, with perception capabilities for track 1, track 2 and track 3. They provide three-dimensional point clouds in a Cartesian coordinate system $(x, y, z)$ and perform the obstacle detection task. Fig. \ref{fig:obst_detection} shows a general overview of the obstacle detection flowchart using 3D LiDAR or stereo camera. 
\paragraph{LiDAR-Based Obstacle Detection}
Unlike commercial LiDAR, the simulated sensor offers neither intensity, nor ring data, since it is an implementation based on ray-casting. Therefore, approaches based on compression of rings and virtual scan, such as those presented in \cite{Hata2016} and  \cite{petrovskaya2009} are not enabled to use during the competition.

Due to such a lack of information on point clouds, we decided to use a simple obstacle detection algorithm based on a height map approach, where the $xy$-plane in the front of the car is divided into a $500\times500$ grid of 10cm resolution. Towards the construction of a height map, each point in the point cloud is compared with the other nearest points contained inside of the same grid cell.  The greatest height difference between two points in the grid cell is compared with a threshold- if it is greater than 15cm, the cell is considered occupied (obstacle); otherwise, the cell is considered a cleared area. The occupied area is saved as a point cloud where, for each point, $(x,y)$ are the same coordinates as the center of the cell and $z$ is a value of height. 

The system segments the point cloud containing the height map (occupied area) into instances and infers basic shapes for each obstacle for making decisions on the obstacles found on the road and performing further tasks, such as tracking.

In the next step, the system uses DBSCAN algorithm \cite{khan2014dbscan} to cluster and separate points belonging to different objects (e.g., cars, pedestrians, poles, among others). DBSCAN returns clusters for each object and a bounding box is then fitted by the rotating calipers algorithm implemented in OpenCV, which returns the min rectangle fitted to the cluster in the $xy$-plane. The height of the cluster completes the 3D box around each cluster. 
\begin{figure}[!tb]
	\centering
	\kern-2.5em
    \subfloat[Left images from \newline stereo\_center camera]{
        \begin{tabular}{c}
            \includegraphics[width=1.7in,height=0.85in]{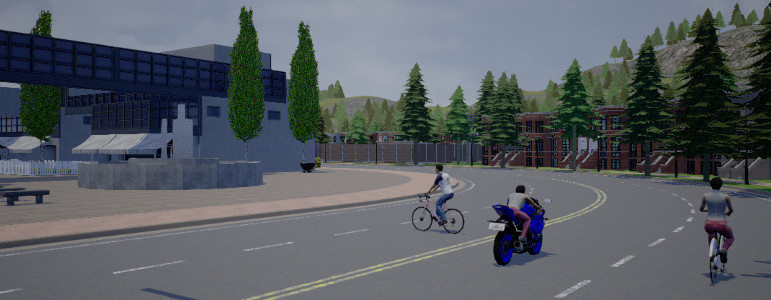}\\
            \includegraphics[width=1.7in,height=0.85in]{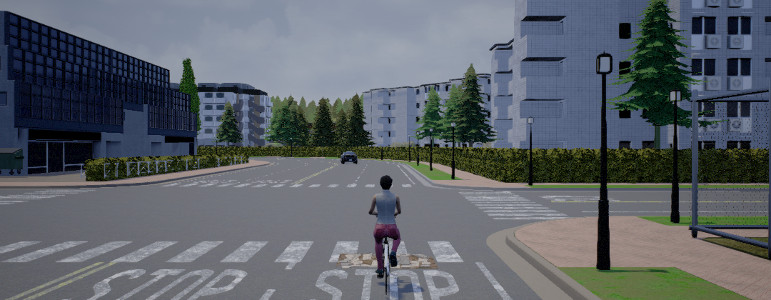}
        \end{tabular}
	}
 	\kern-1.5em
	\subfloat[Obstacles detected on LiDAR \newline point cloud]{
    	\begin{tabular}{l}
    	    \includegraphics[width=1.5in,height=0.85in]{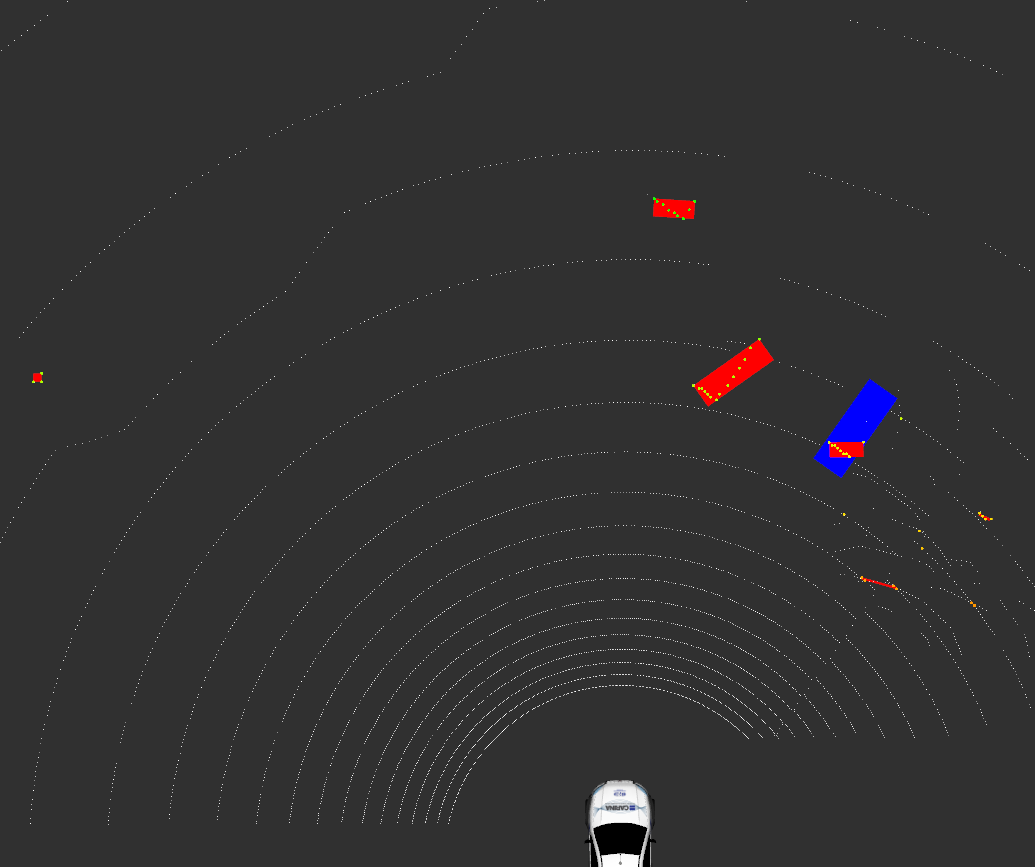}\\
    	    \includegraphics[width=1.5in,height=0.85in]{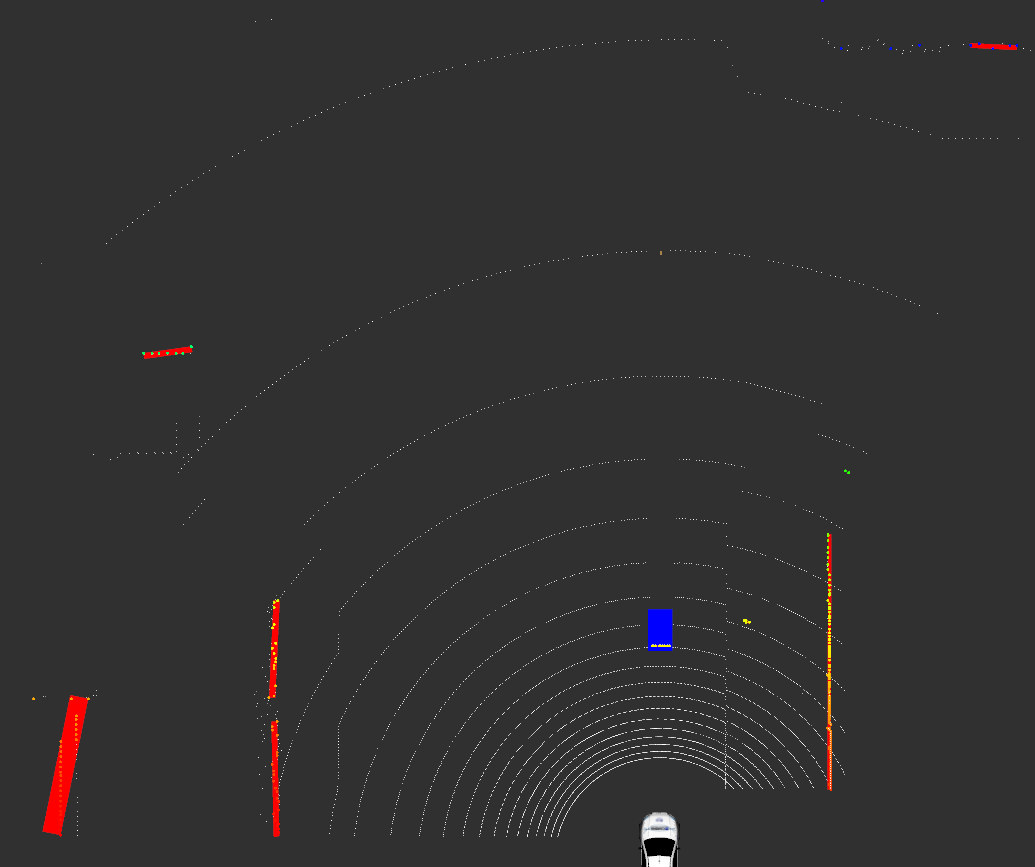}
    	\end{tabular}
	}
	\caption{LiDAR-based obstacle detection}
	\label{fig:obstacle_detection}
\end{figure}

\paragraph{Stereo-based Obstacle Detection}
Driving in urban scenarios requires detailed and precise 3D perception of the environment surrounding the vehicle. LiDAR was an effective sensor to perform this task on \emph{track1} and \emph{track3}. Nevertheless, on \emph{track2}, we are asked to drive to a destination point using cameras only. In the challenge, there are no sensors that provide RGB-D images or depth information further than LiDAR.

We used a stereo system to perform scene reconstruction and obstacle detection. The stereo approach was chosen, since it enables 3D reconstruction with the use of only two monocular cameras with the correct calibration matrix. 

In \emph{track2}, we use stereo cameras to create RGB-D images. The intrinsic and extrinsic parameters of the cameras were manually defined in the autonomous sensor settings.
For \emph{stereo\_left} and \emph{stereo\_right}, we set focal length $f=44$ and field of view $fov=60$ deg, and \emph{stereo\_center} was configured with f=33 and fov=40 deg. In all cases the baseline distance, which is the distance between two cameras in the stereo rig, was 24cm. Short baselines are more accurate for stereo estimation and obstacle detection, and large baselines enable the detection of obstacles far from the agent. A medium baseline was used for balancing these features. Fig. \ref{fig:car_sensing} depicts poses of the three stereo rigs on the car roof. 

ELAS (Efficient LArge-scale Stereo) algorithm extracted disparity maps for depth estimation \cite{geiger2010efficient}, and, subsequently, using disparity maps and camera calibration, the system built depth images and point clouds similar to those generated by LiDAR, but enhanced with RGB information. 

The same algorithm used for obstacle detection using LiDAR's point cloud was applied for obstacle detection using the RGB point cloud from the stereo system. Overall, for \emph{track2}, the RGB point cloud provided the localization of obstacles, traffic lights and traffic signs.

A method based on virtual scan from a 3D LiDAR proposed by Petrovskaya et al. \cite{petrovskaya2009} detected obstacles with the use of a stereo system. Petrovskaya et al. \cite{petrovskaya2009} consider obstacles above the ground and up to two-meter height and build a 3D grid in spherical coordinates for the classification of obstacles into different types. This method was also able to decrease the amount of data in the point cloud. Each spherical grid cell (cone from the origin to the obstacle point) is called virtual ray of the virtual scan. Based on this idea, we used a depth image calculated by the ELAS algorithm considered the depth image a grid in a spherical coordinate system. In this case, the sizes of both spherical grid and columns in the depth image were the same. Columns are grid cells (virtual rays) and rows were compared to rings in a 3D LiDAR. Angles between points contained in rays and angles between rays can be calculated according to the camera field of view and image resolution.

Virtual scan method was applied for the removal of ground and detection of obstacle, as suggested in Petrovskaya et al. \cite{petrovskaya2009}. The first step is to determine obstacle points by circulating the rays (in our case, through the columns in the depth image) from the lowest vertical angle (bottom row) to the highest (top row). In Fig. 7, considering the three points (A, B and C) in the virtual ray, if the two normalized vectors AB and BC are parallel to each other, the angle between them must be close to 0, and the dot product must be close to 1. For perpendicular vectors BC and CD (obstacle), the dot product between them was close to 0. The method eliminated outliers for vectors DE and EF (E point is an outlier) - despite their opposite directions, they are approximately parallel (see Fig. 7) and the dot product will also be close to 1, as in the first case. Therefore, the application of a threshold to the dot product is sufficient for the concomitant elimination of ground and outliers in the depth image. 

This filter aims at eliminating both the spray noise (outliers) caused by an erroneous calculation of disparity in the depth image and flat areas belonging to the ground - only points of obstacles must be maintained. The first objective was successfully achieved, since the spray noise from erroneous disparities was eliminated (see Figs. \ref{fig:stereo_filter_b} and \ref{fig:stereo_filter_d}). The filter worked partially in the second task - while the furthest points belonging to the ground were effectively eliminated, the closest points of the ground were kept, probably due to the proximity between pixels in the closest areas (in this case, vectors AB and BC were not completely parallel and the slope between them might be greater than zero). Fig. \ref{fig:stereo_filter} shows the point cloud generated from the depth image with and without the filter. 

Since the ground had not been completely removed and the obstacles had not been totally separated, the same methods used for 3D LiDAR object detection (height map, clustering) were applied towards separating the obstacles from the ground (see flow diagram in Fig. \ref{fig:obst_detection}). 

Stereo-based system plays two roles, namely support obstacle detection with RGB-D point cloud and localize traffic signs and traffic lights with depth enhancements after 2D detection (see section \ref{section:tlds}). We used the left camera of \emph{stereo\_center} for obstacle detection and classification in an RGB image. A Convolutional Neural Network for object detection and classification was trained with images collected and labeled by our team in several simulated environments under various weather conditions. Our stereo camera system was able to detect obstacles up to 30m distances using our trained classifier and stereo point cloud. 

  \begin{figure}[!tb]
      \centering
      \includegraphics[scale=0.40]{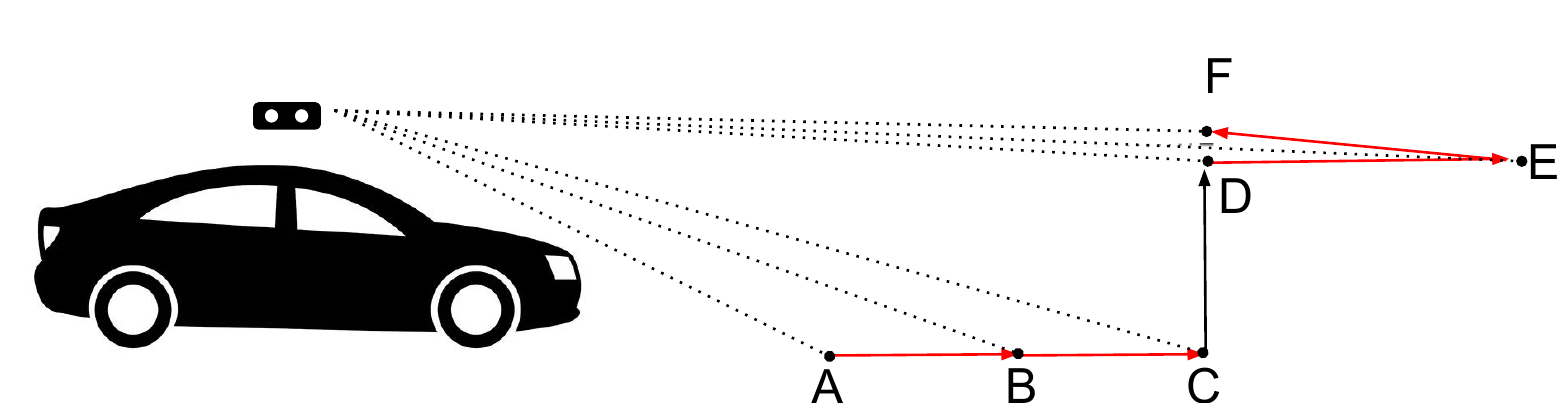}
      \caption{A, B, C, D, E, F represent depth for consecutive pixels in the same column (ray on virtual scan) in depth image. Dot product between AB and BC is approximately 1 (B is a ground point). Same thing for DE and EF (E point is an outlier). In these cases, B and E are rejected.}
      \label{fig:dot_filter}
  \end{figure}

\begin{figure}[!tb]
	\centering
    \subfloat[Frontal View]{
		\includegraphics[width=1.6in]{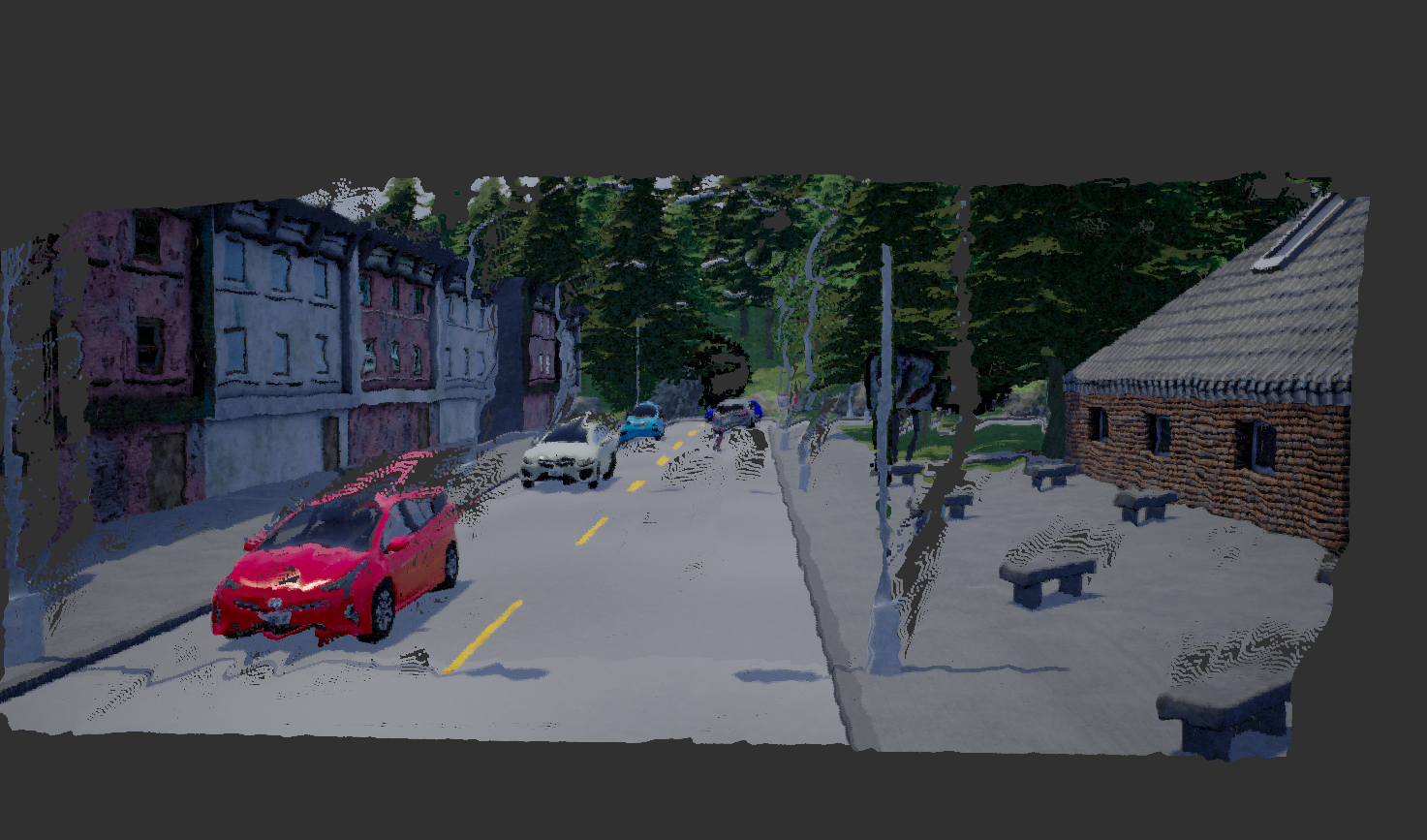}
    	\label{fig:stereo_filter_a}
	}
	\hfil
	\subfloat[Filtered frontal view ]{
		\includegraphics[width=1.6in, height=0.945in]{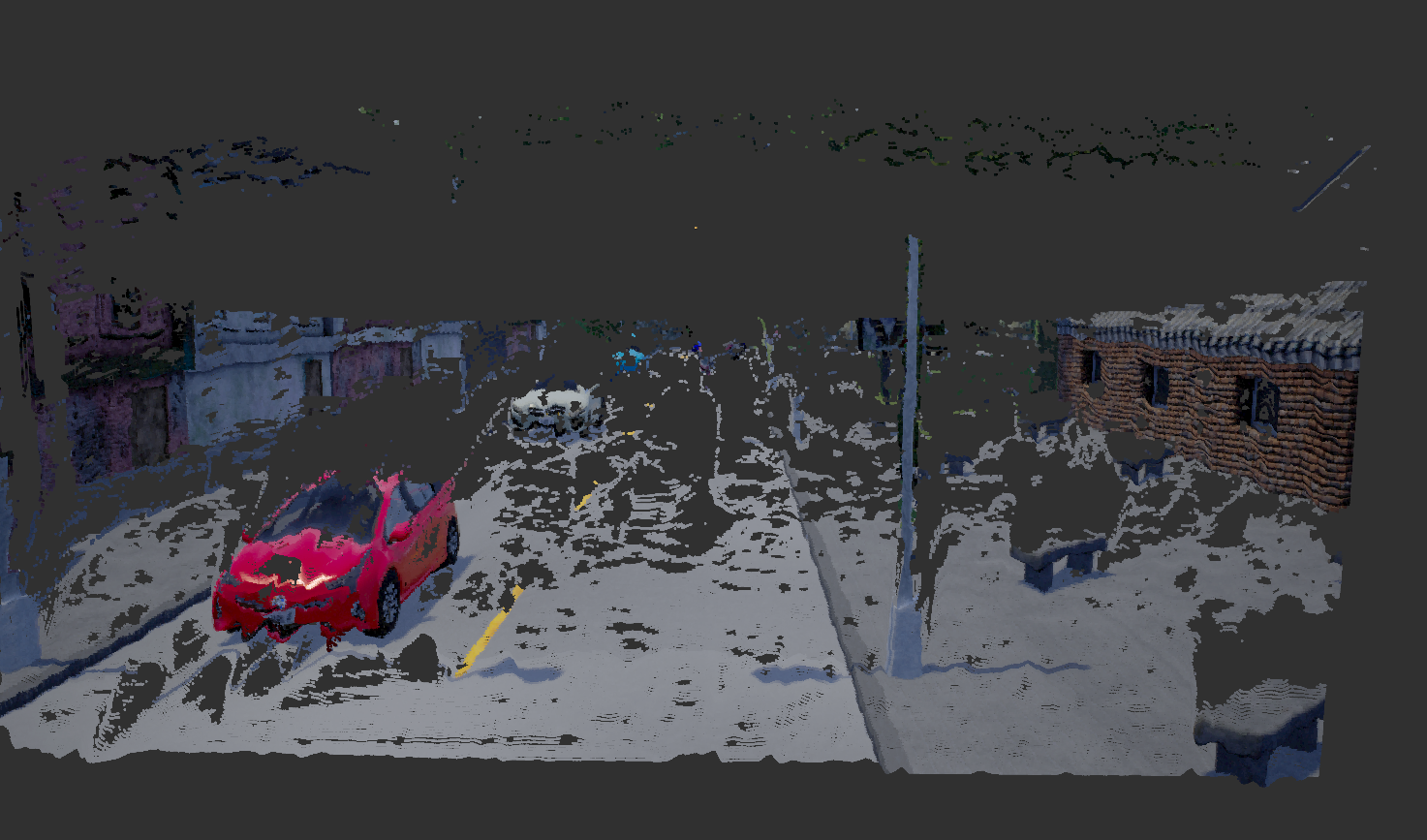}
    	\label{fig:stereo_filter_b}
	}
	\hfil
	\subfloat[Bird eye view]{
		\includegraphics[width=1.6in, height=.87in]{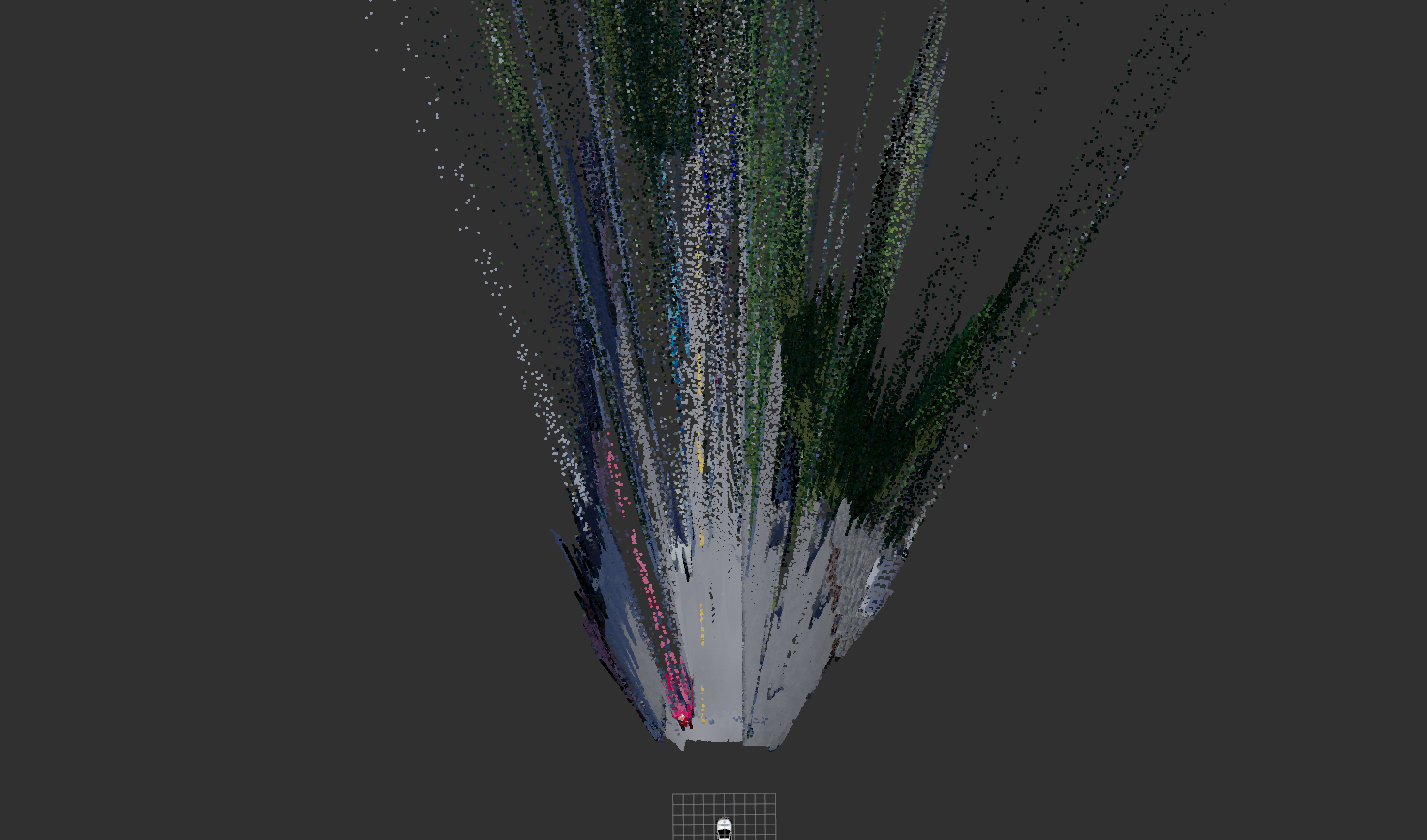}
    	\label{fig:stereo_filter_c}
	}
	\hfil
	\subfloat[Filtered bird eye view]{
		\includegraphics[width=1.6in, height=.87in]{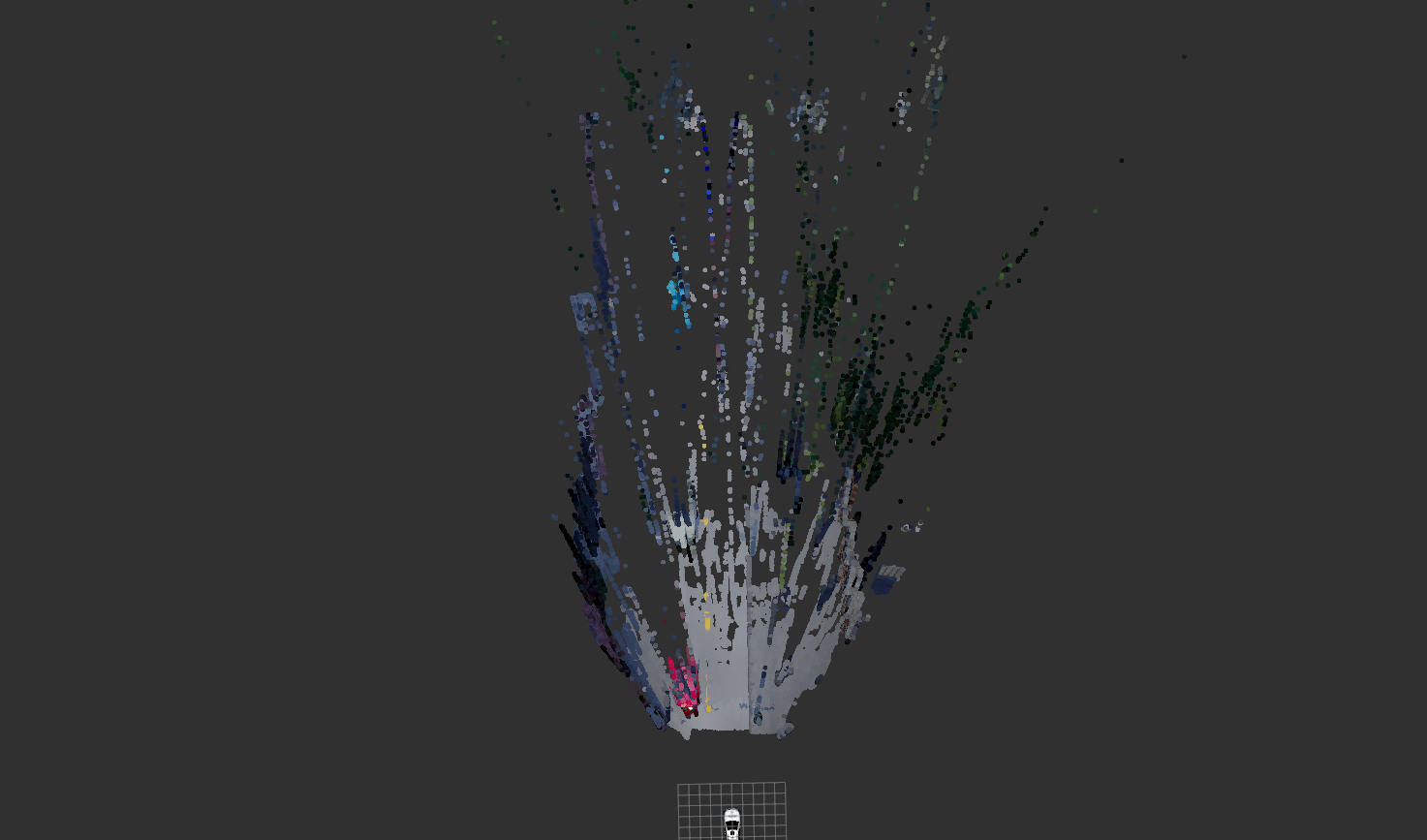}
    	\label{fig:stereo_filter_d}
	}
	\caption{Results for denoising point cloud using dot product between vectors  passing through three adjacent points}
	\label{fig:stereo_filter}
\end{figure}

\subsubsection{Traffic Light and Sign Detection}
\label{section:tlds}
\begin{figure}[!tb]
	\centering
    \subfloat{
		\includegraphics[width=0.15\textwidth]{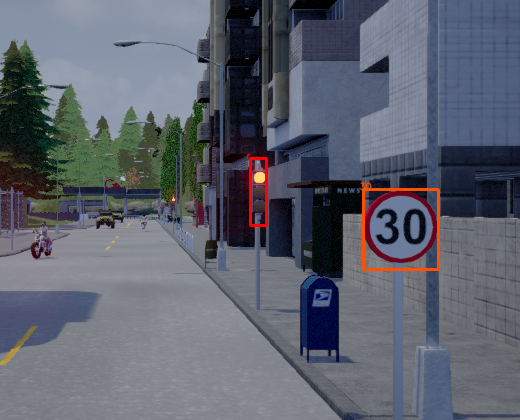}
	}
	\subfloat{
		\includegraphics[width=0.15\textwidth]{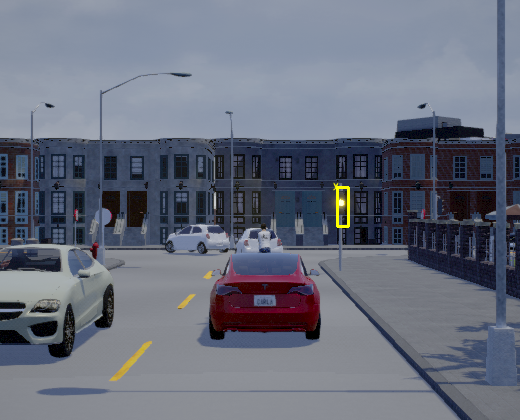}
	}
	\subfloat{
		\includegraphics[width=0.15\textwidth]{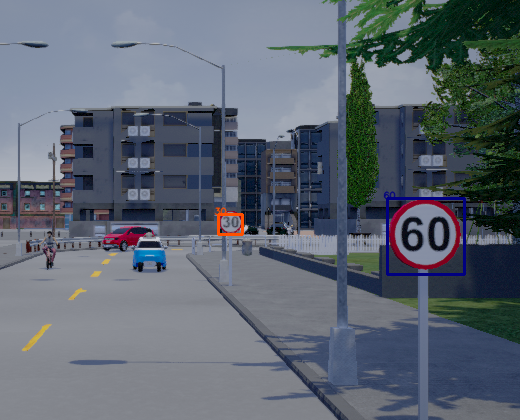}
	}
	\hfil
	\subfloat{
		\includegraphics[width=0.15\textwidth]{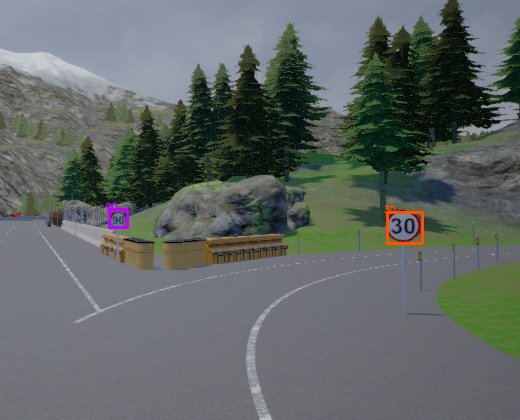}
	}
	\subfloat{
		\includegraphics[width=0.15\textwidth]{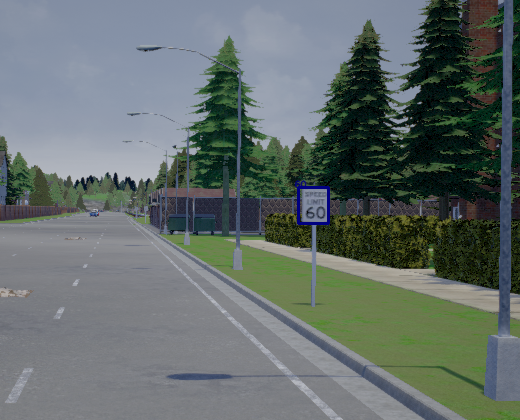}
	}
	\subfloat{
		\includegraphics[width=0.15\textwidth]{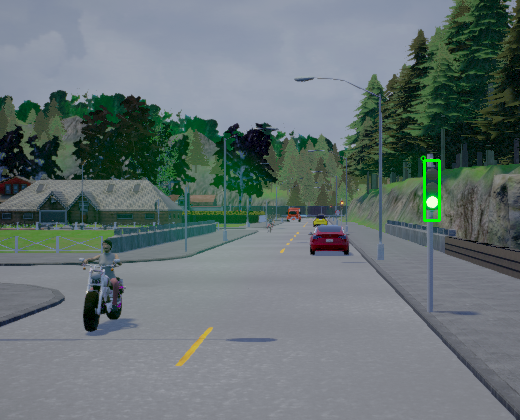}
	}
	\hfil
	\subfloat{
		\includegraphics[width=0.315\textwidth]{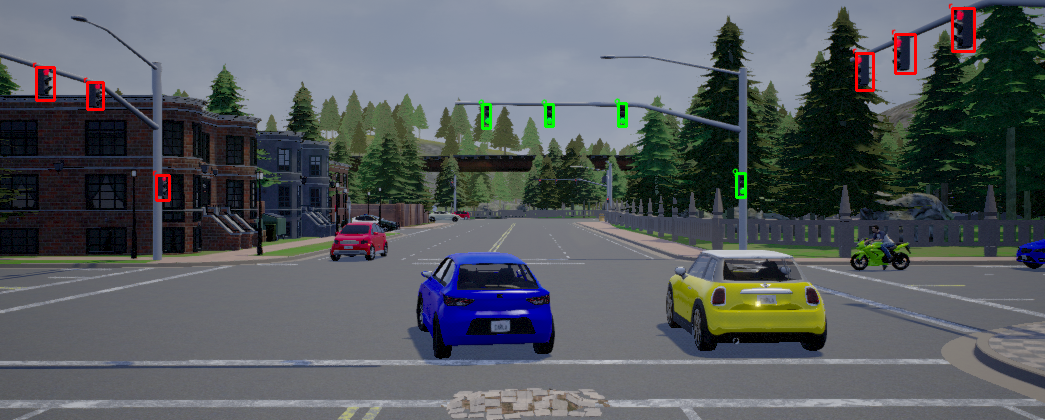}
	}
	\subfloat{
		\includegraphics[width=0.15\textwidth, height=0.89in]{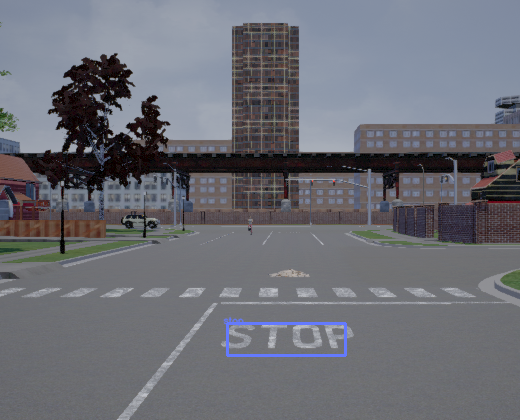}
	}
	\caption{Traffic sign and traffic light detection}
	\label{fig:detection_yolo}
\end{figure}

A Convolutional Neural Network (CNN) known as YOLO v3 was used for the detection of traffic lights and traffic signs in images \cite{yolov3}. An image dataset collected and labeled through the training routes was sliced into training, validation and test sets. The bounding boxes of the annotated images belonged to eight classes, namely horizontal stop, vertical stop, traffic lights (green, yellow, red), and speed plates (90km/h, 60km/h and 30km/h). The original architecture described in Redmon and Farhadi was trained \cite{yolov3}. Fig. \ref{fig:detection_yolo} shows results for inference from the trained detector in different situations on validation routes of CADCH. 
\subsubsection{Hazardous Obstacle Monitor}
Besides obstacle detection, risk assessment is another important component of an autonomous system for ensuring safe driving. According to Hamid et al. \cite{hamid2018review}, it estimates potential threats to accidents, including collision with other traffic participants. Collision is classified into three main types, i.e., collisions with static obstacles, dynamic obstacles, and unexpected obstacles (which may occur due to occlusions). Therefore, both quantitative risk evaluation and binary collision prediction can feed decision-making algorithms towards appropriate actions for the avoidance of collisions. 

According to Kim and Kum \cite{kim2015threat}, and Mechernene et al. \cite{mechernene2018motion}, quantitative risk evaluations use quantitative metrics as risk indicator, such as Time-To-Collision, Time-To-React, Distance-To-Collision, among others. In turn, the binary collision prediction determines whether or not a collision will occur given the current traffic scenario.  Taking into account other traffic participants, Naumann and Stiller \cite{naumann2017towards} proposed a quantitative metric called Time-of-Zone-Clearance (TZC) for estimating the risk of collision between overlapping trajectories. According to the authors, a collision occurs only if paths overlap; therefore, the overlapping zone is a potential collision one. TZC measures the time elapsed between the first vehicle leaving the potential collision zone and the second vehicle entering this area.  

Li et al. \cite{li2015real} developed a path planning algorithm with sampling trajectories, which removes trajectories colliding with obstacles. Artuñedo, Godoy and Villagra \cite{artunedo2019decision} presented a driver corridor path planning and short-term motion prediction for obstacle-aware navigation of autonomous vehicles. Such prediction assumes a constant velocity vector for other traffic participants to reduce computational costs during navigation.  

\begin{figure}[!tb]
      \centering
      \includegraphics[scale=0.40]{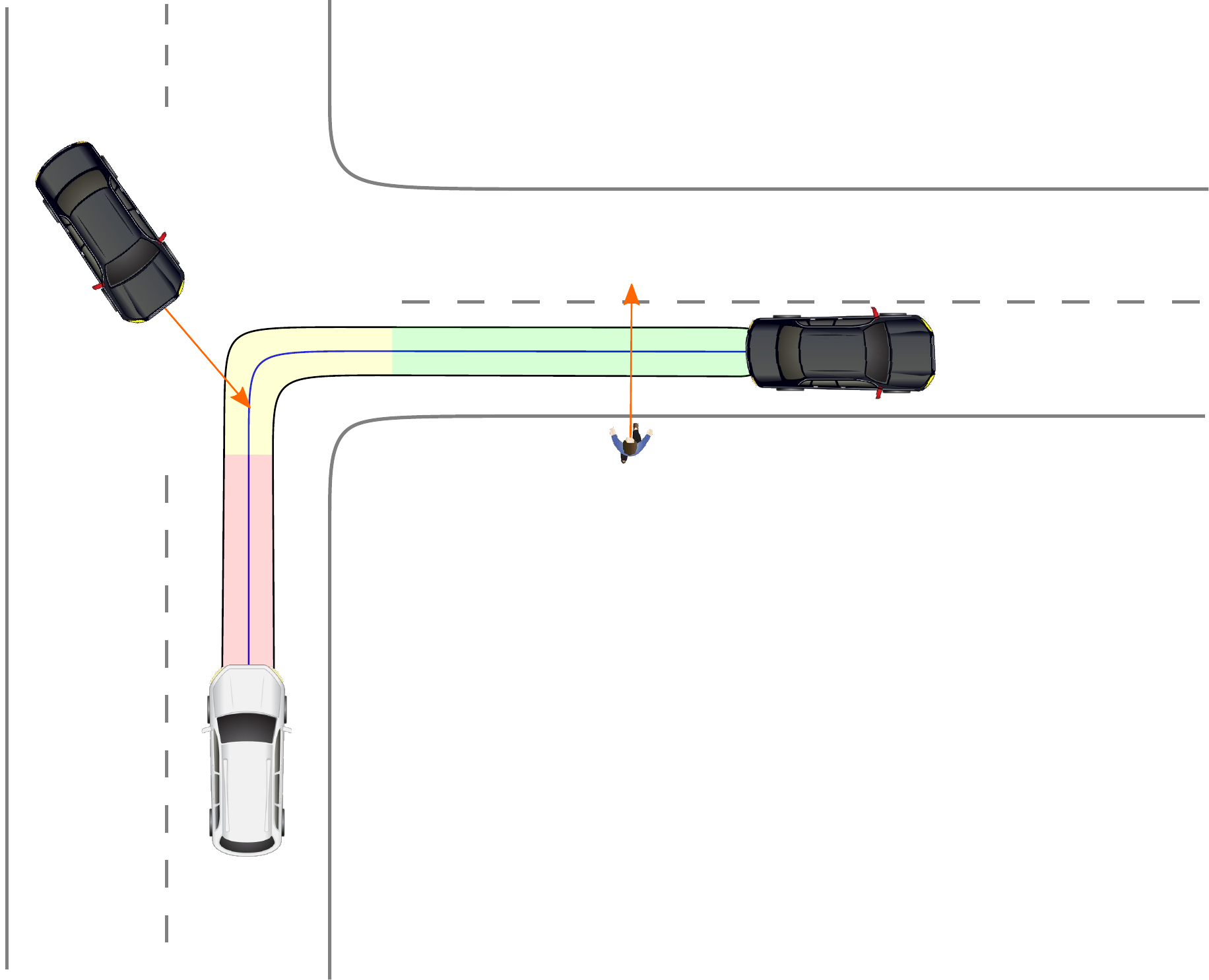}
      \caption{Hazardous Obstacle Monitor}
      \label{fig:obstacle_monitor}
  \end{figure}

The risk assessment component of \emph{CaRINA Agent} creates a safe zone around the ego-vehicle trajectory and applies a short-term motion prediction assuming constant velocity and orientation for estimating the risk of collision between traffic participants. Fig. \ref{fig:obstacle_monitor} illustrates the way the approach works. The ego-vehicle's trajectory is divided into three attention zones regarding its proximity with the ego-vehicle, namely danger, warning, and safe. 
Such zones are used for the classification of different threat levels, improving decision-making on safety actions, i.e., if a trajectory conflict is detected in the first zone (danger), an emergency stop must be executed; however, if it occurs in the 'warning' or 'safe' zones, the vehicle should manage its velocity using Distance-To-Collision (DTC) as input to the decision-making algorithm (see section \ref{subsec:navigation})).  

The detection of conflicts between trajectories assumes two different rules according to the available information on the traffic scene. The first is the position of the obstacles and an estimation of their shape. If their orientation and velocity are available, a short-term motion prediction estimates their future positions assuming constant velocity and orientation, and uses them to detect conflicts. Otherwise, the conflict is determined by the proximity of the obstacle to the ego-vehicle's trajectory with a 1m threshold \cite{hamid2018review}.

%% file: tex/navigation.tex
The navigation system relies only on waypoints to perform each mission, i. e., \emph{CaRINA agent} does not use High Definition Maps or other types of maps to navigate in the simulated urban environment.

\subsubsection{Planning}
The route followed by the agent is composed of waypoints. However, they are generally sparsely positioned between each other, which reduces the smoothness of the path. Towards overcoming this issue, the waypoints are interpolated by cubic splines, so that a dense representation of the path is achieved and smoothness is increased.

\subsubsection{Decision Making}
During navigation, the ego-vehicle must adopt some tactical behaviors for safely reaching its destination (e.g. avoiding static and dynamic obstacles). Considering only static obstacles, the ego-vehicle must plan a trajectory to avoid them, while guaranteeing some level of comfort for the passengers. This task becomes more complex when other traffic participants are involved. For example, intersection negotiation and overtaking require a higher level of reasoning than a simple deviation from a static obstacle on the road. Therefore, the ego-vehicle must consider its future actions as well as state transition of other agents, such as color changes in traffic light. Since communication between the ego-vehicle and infrastructure is not available, the inference of future states of such agents is uncertain. Towards dealing with the inherent uncertainty, the problem can be modeled as a Markov Decision Process (MDP) \cite{brechtel2011probabilistic}.

MDP is a mathematical framework that solves problems subject to uncertainty \cite{Russell2010artificial}. It is defined by the tuple ($\pazocal{S},\; \pazocal{A},\; T,\; R,\; \gamma$), where $\pazocal{S}$ and $\pazocal{A}$ are the state and action space, respectively. When taking action $a \in \pazocal{A}$ in state $s \in \pazocal{S}$, the agent reaches state $s' \in \pazocal{S}$. Conditional probability function $T(s,\; a,\; s')=\text{Pr}(s'|s,\; a)$, which describes the probability of reaching $s'$ from $s$ after taking $a$, models the uncertainty related to state transitions. $R(a,\; s)$ is the expected reward when the action $a$ is taken in $s$. MDP aims to compute a policy $\pi^* : s \to a$ that maximizes the expected sum of discounted rewards:
\begin{equation}\label{eq:policy}
	\pi^*(s) = \underset{\boldsymbol{\pi}}{\text{argmax}} \sum_{t=0}^{\infty} \gamma^t R (s_t,\pi(s_t)),
\end{equation}
where $\gamma \in [0,1)$ is the discount factor, which prioritizes immediate rewards.

The \emph{State space} $\pazocal{S}=[v \quad x_{\varphi} \quad \varphi]^T$ encompasses the velocity $v$ of the agent, the distance $d_{\varphi}$ to the next traffic light and its color $\varphi$, and the distance $d_v$ to a vehicle observed on the agent's route and assumed as a static obstacle.

The agent \emph{action space} is composed of three action: 
\begin{equation}\label{eq:action}\nonumber
    \pazocal{A} = \{Brake, \quad Stay\; Constant, \quad Accelerate\},
\end{equation}
where $Brake$, $Stay\; Constant$ and $Accelerate$ stand for a negative, a null and a positive speed rate change, respectively. The agent's motion is constrained to the following motion model:
\begin{equation}\label{eq:motion_model}
    \begin{bmatrix} v' \\ d_{\varphi}' \\ d_v'  \end{bmatrix} = 
    \begin{bmatrix} v \\ d_{\varphi} \\ d_v \end{bmatrix} + 
    \begin{bmatrix} a_0  \\ -v \\ -v  \end{bmatrix}\Delta t,
\end{equation}
where $a_0$ is the speed rate change of the agent and $\Delta t$ is the time step. The traffic light state $\varphi$ changes according to the stochastic model: 
\begin{equation}\label{eq:tl_model}
    GREEN \xrightarrow{p_1} YELLOW \xrightarrow{p_2} RED \xrightarrow{p_3} GREEN,
\end{equation}
where $p_1$, $p_2$ and $p_3$ represent probabilities of transition.

The \emph{reward model} considers the agent's speed and a reference speed:
\begin{equation}\label{eq:reward_model}
    R(s) = -(v - v_{ref})^2.
\end{equation}
The agent receives a negative reward when its speed deviates from the reference speed, which depends on $d_{\varphi}$, $\varphi$ and $d_v$: if $d_v < 2$ m, or $\varphi = RED$ and $d_{\varphi} < 12$ m, the reference speed is chosen to be null. Otherwise, $v_{ref}$ is chosen to be just below the speed limit of the road. The main parameters of the MDP model are shown in Table \ref{tab:MDPparam}.
\begin{table}[]
\centering
\caption{MDP model parameters.}
\label{tab:MDPparam}
    \begin{tabular}{|l|l|l|l|l|l|l|l|}
    \hline
    $\Delta t$   & $\gamma$ & Brake & StayConst & Acc & $p_1$    & $p_2$   & $p_3$    \\ \hline
    1 s & 0.95  & -4 m/s$^2$    & 0         & 2 m/s$^2$   & 0.05 & 0.6 & 0.05 \\ \hline
    \end{tabular}
\end{table}

Classical methods applied to MDP, such as \emph{value iteration} and \emph{policy iteration} algorithms \cite{Russell2010artificial}, perform poorly in large state spaces, which are assumed to be discrete. Such algorithms also require  $\text{Pr}(s'|s,\; a)$ be explicitly defined. Thereby, an approach that can handle large state spaces while ensuring, at least, near optimal policies must be employed. The online, continuous MDP-POMDP (Partially Observable MDP) solver ABT (\emph{Adaptive Belief Tree}) \cite{Kurniawati2016} is used in the present study.

%% file: tex/control.tex
The \emph{control layer} generates the steering, throttle and brake commands for keeping the agent in the planned trajectory. Decision-making and local path planning modules in the navigation layer set a desired vehicle trajectory in terms of agent action space. 
Two closed control loops are given those reference values and return control actions related to braking, throttle and steering commands, which are directly sent to the simulator interface to be executed. 

\subsubsection{Lateral Control}
The lateral control, which generates the steering signal, is managed by the Model-Based Predictive Control (MPC), in which a cost function is optimized along a predefined time horizon $H$, thus resulting in a sequence of actions, one for each time step $\Delta t$. The immediate action is executed and the process is restarted in the next time step, leading to a receding horizon optimization. 

The formulation of the planner as a convex optimization problem enables the solution computation in a short time step. However, the constraints defining the vehicle's motion model are essentially non-holonomic. Car-like robots can assume positions on the 2-D plane, different headings and steering angles, thus adding up to four degrees of freedom. However, it poses the following two kinematic constraints: a) the vehicle is allowed to move only forward and backward and b) the steering angle is bounded \cite{katrakazas2015real}. Therefore, the actual car motion and the planning trajectory can be different when the planner neglects dynamics factors.

\begin{figure}[!t]
	\centering
	\includegraphics[scale=0.6]{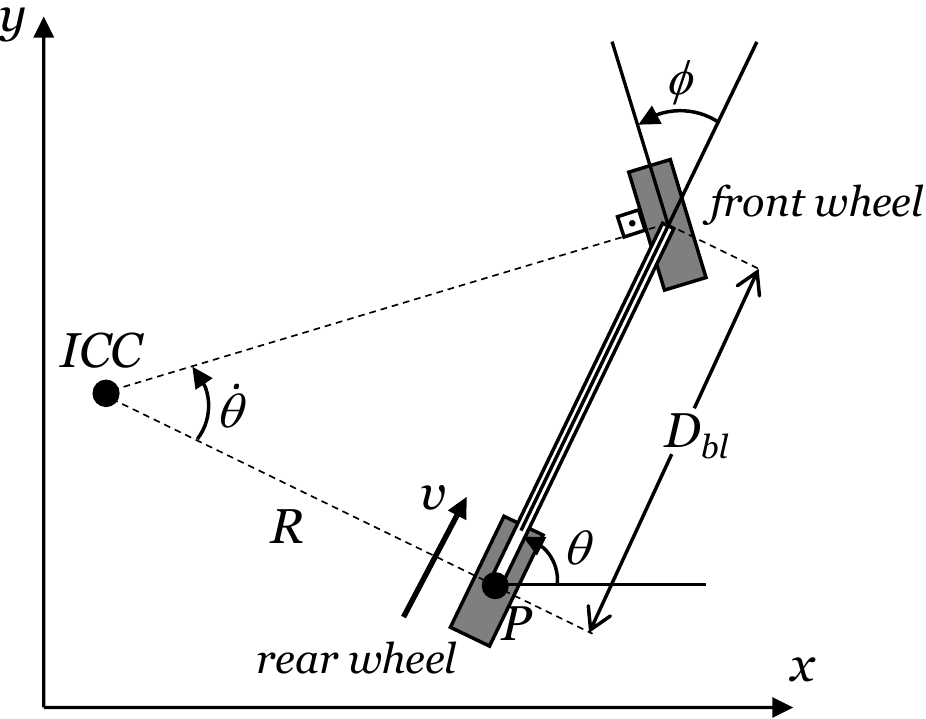}
	\caption{\label{ModBicl} Geometry of a bicycle model. The ICC is determined from the length of the bicycle body $D_{bl}$ and the steering angle $\phi$.}
\end{figure}

Fig. \ref{ModBicl} 
shows a bicycle model used to represent car-like vehicles, which are characterized by Ackerman steering geometry \cite{Dudek_2010}, moving with longitudinal velocity $v$. The front wheel is able to turn and gives the steering angle $\phi$, whereas the rear wheel is always aligned with the bicycle body. According to Fig. \ref{ModBicl}, $\theta$ represents the heading of the vehicle, and $P$ is the guiding point controlled so as to follow the assigned path. The intersection between the lines that pass through the rear and front wheels axes provides the Instantaneous Center of Curvature (ICC). The distance between ICC and $P$ represents the radius of curvature $R$. The curvature of the vehicle is given by $\kappa = 1/R$.

By considering that the wheels roll without slipping, only the kinematic equations can be considered and the lateral dynamic effects can be neglected \cite{Lima_2015}. Therefore, the considerations made so far result in the following kinematic model \cite{Fraichard_2004}
\begin{equation}\label{eq:model}
	\begin{bmatrix} \dot x \\ \dot y \\ \dot \theta \\ \dot \kappa\end{bmatrix} = \begin{bmatrix} \cos{\theta} \\ \sin{\theta} \\ \kappa \\ 0 \end{bmatrix}v + \begin{bmatrix} 0 \\ 0 \\ 0 \\ 1 \end{bmatrix}\tau,
\end{equation}
where $\tau = \dot\phi/(D_{bl}\cos^2{\phi})$.
The motion constraints are added to the optimization problem by means of the third power of $\Delta t$ on the basis of \eqref{eq:model} \cite{obayashi2016appropriate}, where $v$ is computed by the decision-making module (considered constant in the optimization). The cost function is defined as the sum of the quadratic differences between the decision variables and the reference path,
\begin{multline}\label{eq:const1}
	L_{ref} = C_x\frac{1}{2}(x - x_{ref})^2 + C_y\frac{1}{2}(y - y_{ref})^2 + \\+C_{\theta}\frac{1}{2}(\theta - \theta_{ref}) ^2 + C_{\kappa}\frac{1}{2}(\kappa - \kappa_{ref})^2,
\end{multline}
and also, the quadratic of $\tau$
\begin{equation}\label{eq:const2}
	L_{\tau} = C_{\tau}\frac{1}{2}(\tau)^2,
\end{equation}
where $C_x$, $C_y$, $C_{\theta}$, $C_{\kappa}$ and $C_{\tau}$ are cost weights manually tuned. The chosen parameters are shown in Table \ref{tab:MPCparam}. The optimization is performed by \emph{Python} library \emph{scypy.optimize}.
\begin{table}[]
\centering
\caption{Non-linear MPC parameters.}
\label{tab:MPCparam}
    \begin{tabular}{|l|l|l|l|l|l|l|}
    \hline
    $\Delta t$   & $H$   & $C_x$ & $C_y$ & $C_{\theta}$  & $C_{\kappa}$   & $C_{\tau}$  \\ \hline
    1 s & 4 s & 5 & 5 & 10 & 100 & 10 \\ \hline
    \end{tabular}
\end{table}
\subsubsection{Longitudinal Control}
The solution of the MDP problem is the speed rate change to be applied to the agent. Given the current agent's velocity, a new velocity to be tracked can be computed. This tracking is performed by a Proportional-Integral (PI) control.

%% file: tex/results.tex
\subsection{Evaluation Metric}
Autonomous vehicles are heterogeneous and complex systems, composed of a wide range of software components responsible for sensing, perception, decision-making, planning, control, and health-management systems. Different methodologies evaluate such systems. Unit tests aim to analyze each component individually seeking for failures and quantifying its performance according to some metrics (e.g. accuracy, recall and precision for classification algorithms \cite{hossin2015review,tharwat2018classification}). On the other hand, integration tests examine the behavior of two or more components working together (e.g. obstacle detection and obstacle avoidance). Finally, system test evaluates the functioning of the whole system and all its components working concomitantly, and its performance can be summarized by quantitative or qualitative metrics \cite{jorgensen2018software, lewis2017software}. However, no standard methodology assesses and compares the performance of the complete system for autonomous vehicles \cite{hussain2018autonomous}.   
\begin{table}[!tb]
\caption{discounted points due to infractions}
\label{tab:infractions_points}
\centering
\begin{tabular}{|l|c|}
\hline
\textbf{Infractions}                     &\textbf{Discounted Points} \\ \hline
Hitting the static scenery               & 6                       \\ \hline
Hitting another vehicle                  & 6                       \\ \hline
Hitting a pedestrian                     & 9                       \\ \hline
Running a red light                      & 3                       \\ \hline
Invading lane in the opposite direction  & 2                       \\ \hline
Invading a sidewalk                      & 2                       \\ \hline
Running a stop sign                      & 2                       \\ \hline
\end{tabular}
\end{table}

\emph{CARLA Autonomous Driving Challenge} (CADC) proposes a benchmark for the evaluation of autonomous driving systems, which relies on different sets of sensors and software architecture approaches. This competition runs the autonomous system, also known as agent, in a simulated urban environment, where each scenario varies in appearance and architecture of the city, traffic areas (e.g. freeways, urban scenes, residential districts, roundabout, unsigned intersections), size of routes, number of traffic participants, and weather conditions. Moreover, each route is provided with traffic situations based on NHTSA pre-crash typology, such as control loss with no previous action, obstacle avoidance for unexpected obstacles, negotiation at roundabout and unsigned intersection, leading vehicle's break, and crossing intersection with opposite vehicle disobeying traffic lights.   

Towards the evaluation of the agents´ performance in each simulated scenario, the competition proposed a quantitative metric that gathers information on the percentage of each route and infraction committed. It evaluates the performance of the entire system based only on questions (e.g. how well does the autonomous vehicle navigate from a point to a destination obeying traffic rules and ensuring safety of passengers and other traffic participants, while facing common and also unexpected traffic situations(e.g. occluded obstacles and vehicle's control loss.) Equation \ref{eq:cadc_metric} shows the metric, which is an average score of the percentage of complete routes minus the sum of all infractions. Table \ref{tab:infractions_points} displays the discount points for each traffic infraction evaluated. 
\begin{equation}
\label{eq:cadc_metric}
Score(a) = \frac{1}{RN}\sum_{i=1}^{RN} max(100 C(a,r_i)-I(a,r_i),0)
\end{equation}
where $a$ is the agent, $R$ is the number of repetitions for each route, $N$ is the number of routes, $C\left(a, r_i\right)$ is the amount of route completed for the $i$ route, and $I\left(a, r_i\right)$ are the  discounted points in the score due to infractions in each route execution, since $C\left(a, r_i\right) \in \left[0,1\right]$, and $I\left(a, r_i\right) \in \mathbb{N}$. 

\subsection{Challenge and Experimental Results}
To evaluate the \emph{CaRINA Agent}, we present its performance in the competition and also in a simulation experiment using a publicly available set of routes\footnote{These routes are available at \cite{carla_routes}, in the file named \emph{routes\_devtest.xml}.}. The competition involved 10 different routes from unreleased cities' maps, and each of them was executed three times under different weather conditions. On the other hand, the public available set of routes was composed of 26 routes in three different public cities (\emph{Town02}, \emph{Town04}, and \emph{Town05}). 

Our agent was run on an Amazon compute node (AWS) with NVIDIA Tesla K80 GPU (Graphics Processing Unit) and the Robotic Operating System (ROS) kinetic version as the communication middleware. In the simulator experiment, it was run on a computer with Ubuntu 18.04, Intel Core i7-4930K with 32GB memory, NVIDIA GeForce GTX TITAN X video card with 12GB video memory, and the melodic version of ROS.    

Table \ref{tab:results_competition} shows the result of \emph{CaRINA Agent} in each track of the competition. The total score directly reflects the complexity of each track in the competition. The tracks of \emph{Perception Heavy} category (\emph{track1} and \emph{track2}) showed the lowest average scores of complete routes, with $40.82$ (\emph{track1}) and $34.74$ (\emph{track2}), and also a high discount score due to traffic infractions - $16.8$ (\emph{track1}) and $13.43$ (\emph{track2}).

The winner of \emph{track2} \cite{toromanoff2020end} scored $43.45$ of route points, $15.67$ of infractions points and total score of $29.17$. The navigation approach presented in Toromanoff, Wirbel and Moutarde \cite{toromanoff2020end}  uses deep reinforcement learning for vision-based end-to-end urban driving, taking into account tasks such as: lane keeping, traffic light detection, pedestrian and vehicle avoidance, and handling intersection with incoming traffic. This approach outperforms the \emph{CaRINA Agent}, as it navigated a higher average percentage of routes. However, with the availability of more sensors (e.g. LiDAR) in the remaining tracks, \emph{CaRINA 2} surpassed its performance. Their approach achieved $30.46$ of routes points, $5.73$ of infraction points and $24.86$ of total score in \emph{track1}, and $48.93$ of route points, $13.67$ of infraction points and $35.87$ of total score in \emph{track3} \footnote{\url{https://carlachallenge.org/results-challenge-2019/}}.


\begin{table}[!tb]
\caption{Results of the competition}
\label{tab:results_competition}
\centering
\begin{tabular}{|l|c|c|c|}
\hline
\multicolumn{1}{|c|}{\textbf{Tracks}} & \textbf{Route Points} & \textbf{Infraction Points} & \textbf{Total} \\ \hline
track 1                               & 40.82                 & 16.8                       & 26.73          \\ \hline
track 2                               & 34.76                 & 13.43                      & 23.13          \\ \hline
track 3                               & 79.97                 & 13.7                       & 66.83          \\ \hline
track 4                               & 86.58                 & 8                          & 79.12          \\ \hline
\end{tabular}
\end{table}
\emph{Track3} and \emph{track4} in the \emph{Map-based} category achieved the highest score for completing routes, meaning the agents were able to cover most path on each route, and were challenged by more traffic situations. track4 was given the highest score among all tracks, since all perception information was available for the agent, which should be concerned only about decision-making, planning, and control.  

The \emph{CaRINA Agent} scored $0.42$ (\emph{track1}), $0.39$ (\emph{track2}), $0.17$ (\emph{track3}), and $0.09$ (\emph{track4}), by calculating the ratios between the  average infraction scores and the  average percentages of completed route for each track. These performances are in line with the complexity of each track, with the tracks in the \emph{Map-based} category having a considerably lower value than those of the \emph{Perception Heavy}.

Table \ref{tab:results_simulation} shows the results of \emph{CaRINA Agent} for the 26 routes of the simulation experiment. The total score in each track and the average score of complete routes are similar with the results of the competition (Table \ref{tab:results_competition}), which  decreases as complexity of each track increases. The ratios between average infractions score and average percentages of completed route for each track are $0.28$ (\emph{track1}), $0.27$ (\emph{track2}), $0.1$ (\emph{track3}), indicating the agent with fewer sensors and reliable perceptual information committed more infractions due to the inaccuracy representation of the vehicle's surroundings, or even lack of information.  
\begin{table}[!tb]
\caption{Results of the simulation}
\label{tab:results_simulation}
\centering
\begin{tabular}{|l|c|c|c|}
\hline
\multicolumn{1}{|c|}{\textbf{Tracks}} & \textbf{Route Points} & \textbf{Infraction Points} & \textbf{Total} \\ \hline
track 1  & 43.74    & 12.46   & 32.03   \\ \hline
track 2  & 39.33    & 10.76   & 29.15   \\ \hline
track 3  & 76.91    & 7.61    & 69.29   \\ \hline
track 4  & 96.27    & 4.03    &92.23 \\ \hline
\end{tabular}
\end{table}
\subsection{Traffic Infractions}
\begin{figure}[!tb]
	\centering
    \subfloat{
		\includegraphics[width=0.15\textwidth]{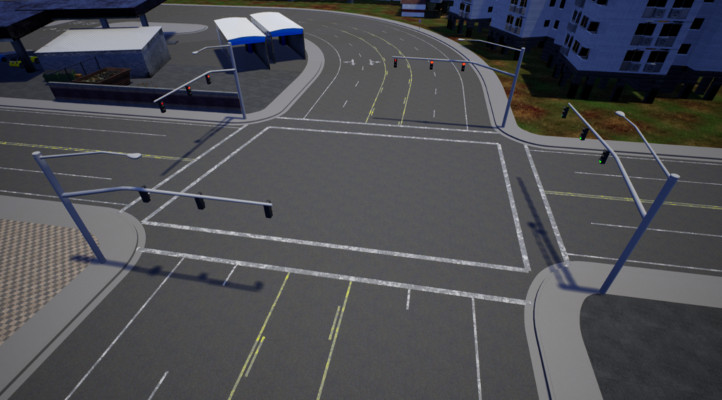}
	}
	\subfloat{
		\includegraphics[width=0.15\textwidth]{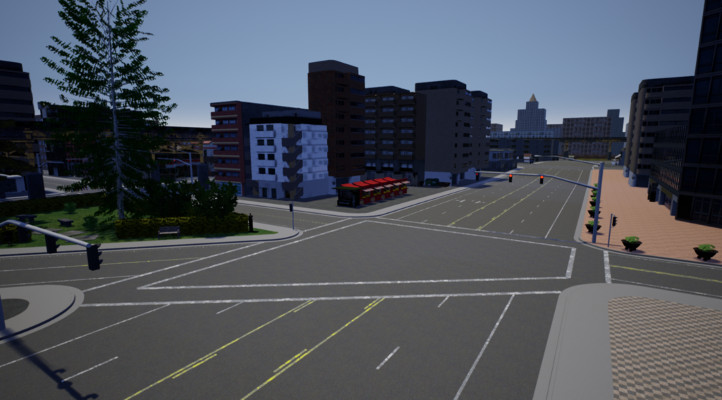}
	}
	\subfloat{
		\includegraphics[width=0.15\textwidth]{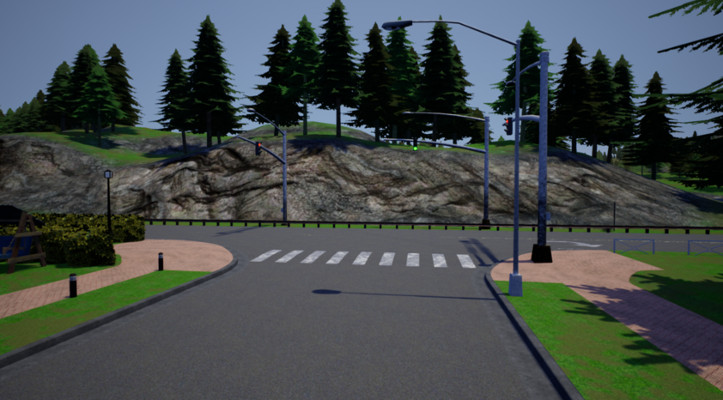}
	}
	\hfil
	\subfloat{
		\includegraphics[width=0.15\textwidth]{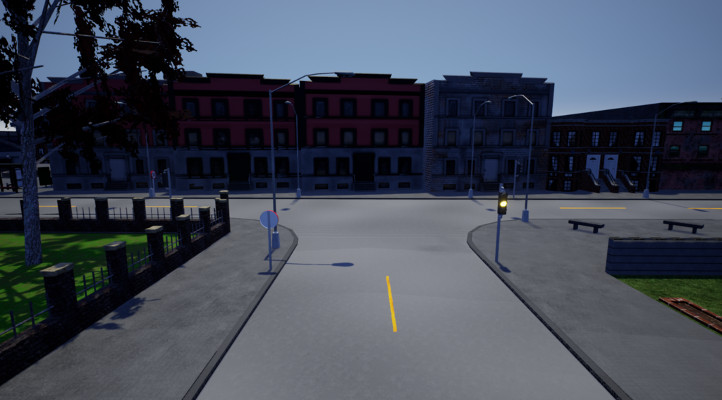}
	}
	\subfloat{
		\includegraphics[width=0.15\textwidth]{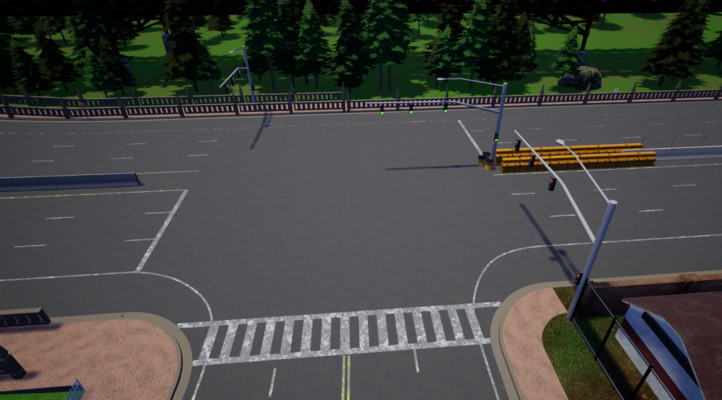}
	}
	\subfloat{
		\includegraphics[width=0.15\textwidth]{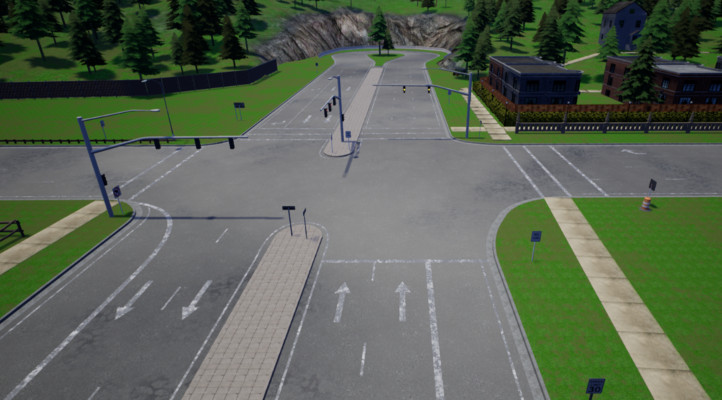}
	}
	\hfil
	\subfloat{
		\includegraphics[width=0.15\textwidth]{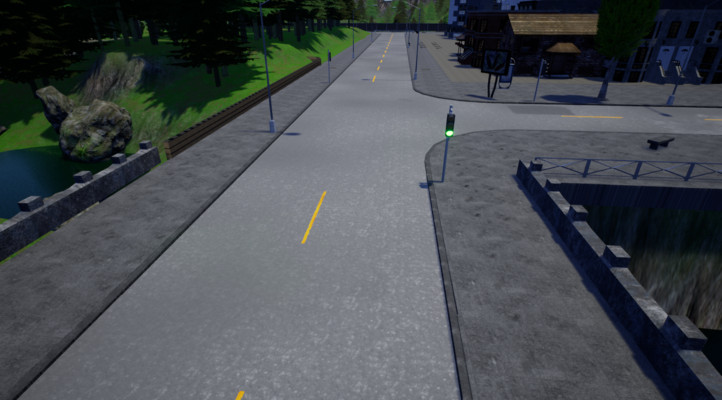}
	}
	\subfloat{
		\includegraphics[width=0.15\textwidth]{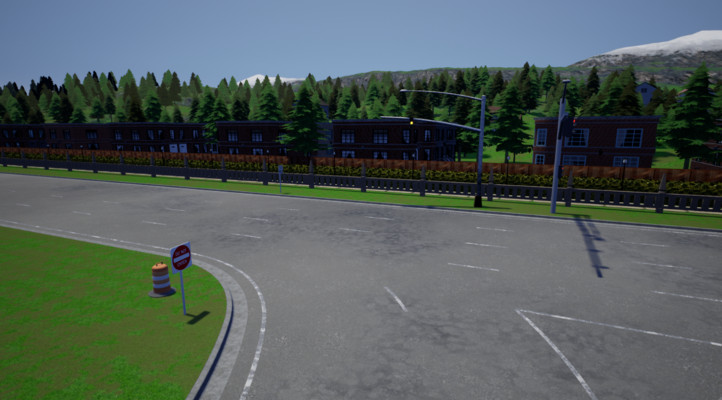}
	}
	\subfloat{
		\includegraphics[width=0.15\textwidth]{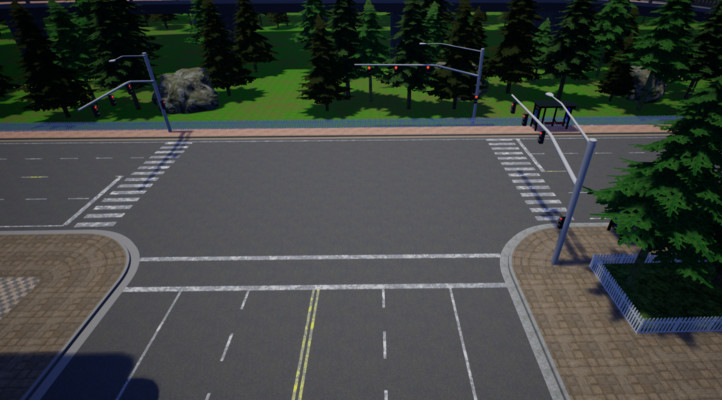}
	}
	\caption{Challenging Traffic Light Scenarios and Junctions' Layout}
	\label{fig:traffic_light_scenarios}
\end{figure}

The lower performance of tracks that used perception algorithms to create the surroundings representations (\emph{track1}, \emph{track2}, and \emph{track3}) can be explained by their challenge in representing different urban scenarios. For instance, CARLA simulator's cities have different designs for traffic light poles and junction's layouts (Fig. \ref{fig:traffic_light_scenarios}). In each case, the autonomous system should identify the correct traffic light to be obeyed and the position at which it should stop. The stop position also depends on the position of the system's sensor, since the vehicle must continue to observe the traffic lights for updating its status.  

The influence due to lack of precision in surrounding representation on the agent's performance is shown in Tables \ref{tab:amout_infractions_competition} and \ref{tab:infractions_simulation}, which provide the number of traffic infractions \emph{CaRINA Agent} committed, respectively, in the competition and in the simulation experiment. The agents of \emph{track1}, \emph{track2}, and \emph{track3} committed more \emph{"Running a red light"} infractions due to the necessity of detecting traffic lights and stop lines. This information is provided in \emph{track4} by the pseudo-sensor called \emph{ObjectFinder}, which justifies the small number of infractions by the agent.  In other autonomous vehicle architectures, High Definition Maps (HD-Map) provide information like this, for example, the Lanelets 2 \cite{poggenhans2018lanelet2}, OpenDriver \cite{dupuis2010opendrive}, and Tsinghua map model \cite{JIANG2019305}. Therefore, maps with augmented semantic and traffic information can improve the performance of autonomous systems, if used in conjunction with perceptual information for increasing accuracy and reliability of the vehicle's surroundings representation and understanding \cite{zang2018high}. 

Another challenging scenario for autonomous systems that heavily rely on perception algorithms is to recognize the navigable area of environments. In \emph{track1} and \emph{track2}, the agents were provided with only a sparse representation of the route, with waypoints and high-level commands, i.e. turn left, turn right, go straight, change lane left, and change lane right. We employed an interpolation approach using the sparse waypoints towards creating a denser representation of the trajectory. However, the diversity in the roads' geometry layout is still an important issue to be considered. Therefore, the agents of \emph{track1} and \emph{track2} committed several infractions of "Invading lane in opposite direction" and "Invading a sidewalk" types (see Tables  \ref{tab:amout_infractions_competition} and \ref{tab:infractions_simulation}), which highlights the need for more reliable representations of the vehicle's path and navigable surroundings area, such as road segmentation that use images and point clouds \cite{li2013sensor,shinzato2014road,mendes2016exploiting,meyer2018deep}, and road boundary detection \cite{zhang2018road, sun20193d}.
\begin{table}[!tb]
\caption{Amount of infractions during the competition}
\label{tab:amout_infractions_competition}
\begin{center}
\begin{tabular}{|c|c|c|c|c|}
\hline
\textbf{Infractions} & \textbf{Track 1} & \textbf{Track 2} & \textbf{Track 3} & \textbf{Track 4} \\
\hline
\begin{tabular}[c]{@{}c@{}}Hitting the\\static scenery\end{tabular}     & 3       & 3       & 0       & 8       \\ 
\hline
\begin{tabular}[c]{@{}c@{}}Hitting a\\ pedestrian\end{tabular} & 20      & 10      & 23      & 10      \\ 
\hline
\begin{tabular}[c]{@{}c@{}}Hitting another\\ vehicle\end{tabular}   & 28      & 38      & 17      & 12      \\ 
\hline
\begin{tabular}[c]{@{}c@{}}Invading lane in\\ the opposite\\ direction\end{tabular}        & 32      & 12      & 0       & 0       \\ 
\hline
\begin{tabular}[c]{@{}c@{}}Invading a\\ sidewalk\end{tabular}   & 4       & 2       & 0       & 0       \\ 
\hline
\begin{tabular}[c]{@{}c@{}}Running a\\ red light\end{tabular}   & 22      & 13      & 34      & 10      \\ 
\hline
\end{tabular}
\end{center}
\end{table}
\begin{table}[!tb]
\caption{Amount of infractions during the simulation}
\label{tab:infractions_simulation}
\begin{center}
\begin{tabular}{|c|c|c|c|c|}
\hline
\textbf{Infractions}  & \textbf{Track 1} & \textbf{Track 2} & \textbf{Track 3} & \textbf{Track 4} \\
\hline
\begin{tabular}[c]{@{}c@{}}Hitting the\\static scenery\end{tabular}                    & 15       & 12      & 0       & 0       \\ 
\hline
\begin{tabular}[c]{@{}c@{}}Hitting another\\ vehicle\end{tabular}                      & 18       & 21      & 15      & 7      \\ 
\hline
\begin{tabular}[c]{@{}c@{}}Invading lane in\\ the opposite\\ direction\end{tabular}      & 16       & 7       & 0       & 1       \\ 
\hline
\begin{tabular}[c]{@{}c@{}}Invading a\\ sidewalk\end{tabular}    & 11       & 10       & 4       & 5       \\ 
\hline
\begin{tabular}[c]{@{}c@{}}Running a\\ red light\end{tabular}    & 16       &12        & 28      & 15      \\ 
\hline
\begin{tabular}[c]{@{}c@{}}Running a\\ stop sign\end{tabular}    & 12       & 6        & 8      & 3       \\ 
\hline
\end{tabular}
\end{center}
\end{table}
Besides path tracking and traffic rules compliance, the autonomous system must avoid collision with obstacles for ensuring safety of its passengers and other traffic participants. It must detect the obstacle, understand the traffic scenario, assess the risk associated with the situation, make decisions, and perform the action for avoiding the collision, which may involve changing its trajectory or executing an emergency braking. Some of the challenges related to obstacle avoidance are different traffic scenarios such as highways, intersections and roundabout, which may require negotiation among traffic participants, field of view of the sensors, which may create blind spots around the vehicle, and other traffic participants' behaviors, which may lead to unexpected and dangerous situations.  

We adopted a point cloud clustering approach for obstacle detection, and a corridor-based approach to assess collision risk. The latter creates a safety-field around the vehicle's path for estimating the likelihood of collisions with obstacles near the vehicle. However, collision was the most frequent traffic infraction of \emph{CaRINA Agent} in both competition and simulated experiment, with ratios of collision by percentage of completed route of $1.25$ (\emph{track1}), $1.47$ (\emph{track2}), $0.5$ (\emph{track3}), $0.35$ (\emph{track4}) in the competition, and  $0.75$ (\emph{track1}), $0.84$ (\emph{track2}), $0.2$ (\emph{track3}), $0.07$ (\emph{track4}) in the simulated experiment.

Collisions with static obstacles are related to sidewalk invasion, and collision with pedestrians with the occluded and unexpected obstacles that appear close to the vehicle, not giving it enough time to avoid the collisions. This reinforces the need to have routines available for carrying out emergency maneuvers, which are executed in a shorter time interval, even if this leads to more sudden braking \cite{rosen2013autonomous,schraner2019pedestrian}, and also pedestrian's behavior anticipation and trajectory prediction \cite{ridel2018literature,ridel2019understanding, ridel2020scene, li2019pedestrian, bighashdel2019survey}.

Collisions with other vehicles occur mainly at intersections and when the vehicle ahead decelerates or stops, and the \emph{CaRINA Agent} does not react in time to avoid a collision. In the first case, the number of collisions can be reduced by integrating components for decision-making in the system's architecture. Such components take into account vehicles' negotiation, which can be provided by vehicle-to-vehicle communication \cite{chen2015cooperative,hafner2011automated,de2017traffic}, or through the estimation of the intention and trajectory of other traffic participants to find the best time to cross an intersection \cite{tran2014online,tawari2016predicting,roy2019vehicle,noh2018decision}. Similarly, the latter situation can also be avoided by vehicle-to-vehicle communication \cite{milanes2011making}, i.e., the vehicle ahead advises other vehicles on its action, or by predicting its intention according to its actions over a period of time, which provides the ego-vehicle with enough time to react \cite{lefevre2014survey,xin2018intention,deo2018would,deo2018convolutional}.      

%% file: tex/conclusion.tex
This article has addressed the development of a software architecture for an autonomous vehicle designed towards meeting the CARLA Autonomous Driving Challenge (CADC). The vehicle used GPS, cameras and LiDAR for a reliable perception system, together with a planning and decision-making system based on Markov Decision Process for completing several routes in many traffic scenarios autonomously. 

LRM-B team successfully completed the challenge and ranked first position in tracks 1, 3 and 4, and second in track 2. \emph{CaRINA Agent} was able to drive autonomously for several kilometers, obeying basic and complex traffic laws, negotiating stop and traffic light intersections, and merging into simulated traffic according to methods and algorithms well-known in robotics and in the autonomous driving research field. Such methods and algorithms include obstacle detection using height maps, clustering, stereo vision, deep learning, path planers, motion controllers, extended Kalman filters for localization, decision-making, and a hazardous obstacle monitor.  

The system developed by the LRM-B team for CADC proved robust and extensible. Initially, the approach was focused on \emph{track3}; however, the architecture was soon extended towards meeting the requirements of the remaining tracks, since additional capabilities could be added or easily changed due to the modular architecture. The success of LRM-B Team was largely based on previous experience with our experimental platforms (\emph{CaRINA 1},  \emph{CaRINA 2} and \emph{Smart Truck}) \cite{Fernandes2014}.

During the challenge, among the several novel aspects developed and tested is the use of short-term motion prediction for risk assessment and Distance-to-Collision for decision-making. Another contribution was the decision-making approach for longitudinal high-level control by a probabilistic framework, known as Markov Decision Process, which handled different traffic scenarios, such as traffic jams, highways, and signalized intersections. The use of 2D object detection and projection in the 3D point cloud (stereo and LiDAR) for finding the 3D poses of obstacles, traffic lights and signs must also be highlighted, since it improved decision-making. 

Although successful in CADC, our architecture and methods should be extended for a real autonomous drive vehicle including other software components to improve risk assessment such as a  3D  obstacle  detection  with  shape  and orientation estimation, multi-object tracking, intention prediction and long-term trajectory prediction of traffic participants. Improved software components related to decision-making, path planning and road boundary segmentation can be added to reduce the lane sidewalk invasion. Besides that, a decision-making component can be used for unsigned intersections.

As future work, we aim at using visual navigation approaches for tracks with no prior knowledge of the map, (e.g., tracks 1 and 2 of the competition). Some of such approaches are imitation learning and end-to-end deep learning, which apply machine learning techniques to directly generate the control commands of a vehicle, i.e. brake, throttle and steering \cite{pan2017agile,codevilla2018end,liang2018cirl,amado2019end,huang2020multi}. 

Finally, the architecture developed provides an interesting platform for future research in all fields of autonomous driving, and a starting point for more complex and advanced architectures. CARLA simulator has proven an interesting tool that, together with its development framework, can accelerate both prototyping and testing of autonomous vehicles in a relatively short time, and provides a benchmark for autonomous systems evaluation. 